\documentclass[letterpaper,journal]{IEEEtran}
\usepackage{amsmath,amsfonts,amssymb}
\usepackage{array}
\usepackage{booktabs}
\usepackage{multirow}
\usepackage{graphicx}
\usepackage{adjustbox}
\usepackage[table]{xcolor}
\usepackage{cite}
\usepackage{url}
\usepackage[colorlinks=true,linkcolor=red,citecolor=blue,urlcolor=blue]{hyperref}
\makeatletter
\renewcommand{\@cite}[2]{\textcolor{blue}{[{#1\if@tempswa , #2\fi}]}}
\let\BETA@oldeqref\eqref
\renewcommand{\eqref}[1]{\textcolor{red}{\BETA@oldeqref{#1}}}
\makeatother
\graphicspath{{./}}
\definecolor{diag_green}{RGB}{232,245,233}
\definecolor{avg_yellow}{RGB}{255,243,224}
\definecolor{last_red}{RGB}{248,232,232}
\definecolor{trans_purple}{RGB}{232,234,246}
\definecolor{gain}{RGB}{0,128,0}
\definecolor{loss}{RGB}{0,80,160}
\begin{document}
\title{Black-Box Continual Learning for Vision-Language Models}
\author{Yuting Li,
        Weihang Fang,
        Haoyuan Gao,
        Linghe Kong,
        Yexin Li,
        Lichao Sun,
        and Weiran Huang
\thanks{Yuting Li is with the School of Computer Science, Shanghai
Jiao Tong University, Shanghai, China (e-mail: yutingli@sjtu.edu.cn).}
\thanks{Weihang Fang is with the School of Automation and Intelligent
Sensing, Shanghai Jiao Tong University, Shanghai, China (e-mail:
scgs402@sjtu.edu.cn).}
\thanks{Haoyuan Gao is with the School of Computer Science, Shanghai
Jiao Tong University, Shanghai, China (e-mail: gaohaoyuan@sjtu.edu.cn).}
\thanks{Linghe Kong is with the School of Computer Science, Shanghai
Jiao Tong University, Shanghai, China (e-mail: linghe.kong@sjtu.edu.cn).}
\thanks{Yexin Li is with the State Key Laboratory of General Artificial
Intelligence, Beijing Institute for General Artificial Intelligence
(BIGAI), Beijing, China (e-mail: yliby@connect.ust.hk).}
\thanks{Lichao Sun is with the Department of Computer Science and
Engineering, Lehigh University, Bethlehem, PA, USA (e-mail:
lis221@lehigh.edu).}
\thanks{Weiran Huang is with the School of Computer Science, Shanghai
Jiao Tong University, and Shanghai Innovation Institute, Shanghai,
China (e-mail: weiran.huang@outlook.com).}}
\markboth{SUBMITTED TO IEEE TRANSACTIONS ON PATTERN ANALYSIS AND MACHINE INTELLIGENCE}
{SUBMITTED TO IEEE TRANSACTIONS ON PATTERN ANALYSIS AND MACHINE INTELLIGENCE}
\maketitle
\begin{abstract}
The rapid deployment of Vision-Language Models (VLMs) in dynamic environments necessitates the ability to learn continuously without forgetting. However, traditional continual learning (CL) settings often rely on white-box paradigms, which is increasingly invalidated by the shift toward cloud-hosted models. In this paper, we introduce Black-CL, a more realistic benchmark for VLMs that enforces three primary real-world challenges: weight and architecture inaccessibility, constrained computation, and task-agnostic inference. The learner can query only output embeddings or logits, with no gradient flow through or structural modification of the backbone. Current CL methodologies, which rely on backbone backpropagation or complex parameter expansion, are fundamentally incompatible with these constraints. Under this setting, we propose BETA, a simple yet effective baseline built on the key insight that solely optimizing textual prototypes can navigate the complexities of CL. BETA integrates three core components: Semantic Projection Accumulation (SPA) for incremental knowledge acquisition, Latent Distribution Replay (LDR) for anchoring the embedding space against catastrophic forgetting, and Test-Time Prototype Adaptation (TTPA) for dynamic, instance-aware boundary refinement. Extensive experiments across ten diverse datasets and various backbones demonstrate that BETA significantly outperforms existing black-box tuners. Remarkably, with only 0.05 M trainable parameters, a 180--3000$\times$ reduction compared to competitive methods, BETA achieves performance on par with or even exceeding white-box CL methods. We believe Black-CL and BETA provide a foundational framework for future advancements in continual learning and accelerates the transition of continual learning from academia to real-world systems.
\end{abstract}
\begin{figure}[!t]
    \centering
    \includegraphics[width=\columnwidth]{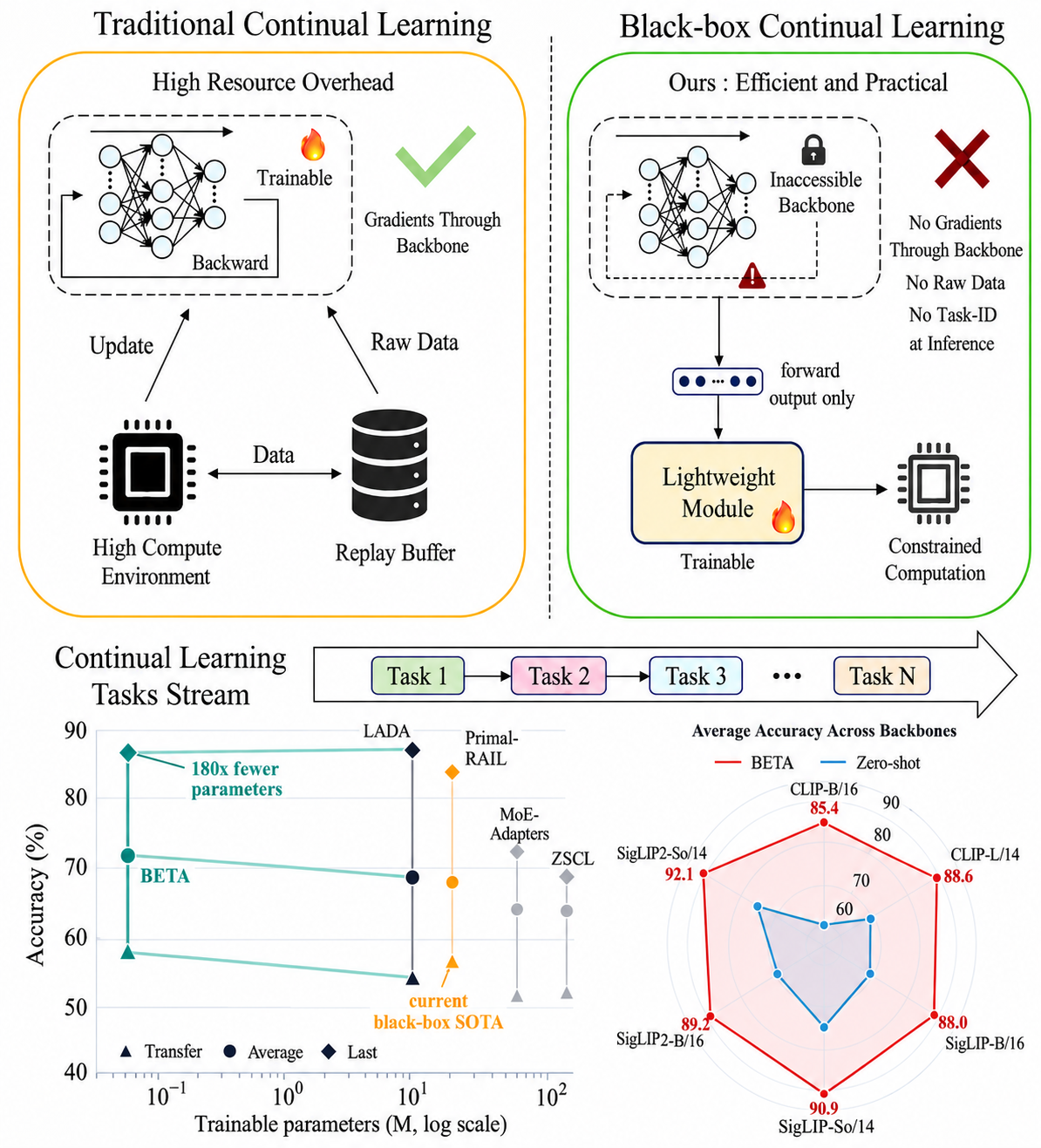}
    \caption{\textbf{Comparison between traditional continual learning (CL)
    and the proposed Black-CL setting, with a white-box SOTA comparison.}
    Conventional CL assumes white-box
    access to the backbone, allowing gradients to propagate through the model
    and permitting internal modifications such as adapters. Black-CL exposes
    only model outputs, such as embeddings or logits: the backbone is not
    merely frozen, but its parameters, architecture, intermediate states, and
    backward graph are inaccessible. Consequently, adaptation must be
    performed entirely outside the backbone without gradient flow through it;
    BETA satisfies this constraint by learning a lightweight prototype
    classifier solely from queried output embeddings. BETA uses very few
    trainable parameters while surpassing the existing black-box SOTA method,
    and it is closely comparable to white-box SOTA approaches, even exceeding
    them on several metrics. The bottom-right panel further shows that the
    improvements brought by BETA are consistent across different VLM
    backbones.}
    \label{fig:sliding_window}
\end{figure}
\section{Introduction}
\IEEEPARstart{T}{HE} remarkable success of Vision-Language Models (VLMs), such as CLIP~\cite{radford2021learning}, has revolutionized open-vocabulary recognition across various downstream tasks~\cite{liang2023open, wu2023cora, yu2023convolutions, bai2025self}. In real-world applications, these models are increasingly deployed in dynamic environments where data arrives in a continuous stream. This setup requires the system to learn new knowledge without forgetting previously acquired knowledge~\cite{rebuffi2017icarl, hsu2018re}, a challenge that has fueled the development of Continual Learning (CL) for VLMs~\cite{zheng2023preventing, yu2024boosting, xu2024advancing, luo2025ladascalablelabelspecificclip}. However, despite substantial progress, a significant gap remains between existing academic benchmarks and the practical requirements of real-world deployment.
Traditional CL paradigms predominantly operate under a "white-box" assumption that grants researchers full access to model weights and gradients. While effective in controlled settings, this assumption is increasingly invalidated by the modern AI landscape. Many state-of-the-art foundation models~\cite{achiam2023gpt, team2023gemini} are now hosted as proprietary services or deployed on edge devices, meaning the backbone architecture and weights remain entirely inaccessible~\cite{sun2022black, sun2022bbtv2, diao2022black, oh2023blackvip, ouali2023black}. Furthermore, practical CL requires models to perform task-agnostic inference by discriminating among all seen classes in a global label space without a task oracle. These models must also operate under the constrained computation typical of edge hardware. 
To bridge this gap, we introduce Black-Box Continual Learning (Black-CL) as a more realistic benchmark for VLMs. In this work, the term ``black-box'' follows its usage in black-box transfer learning and tuning~\cite{guo2023black,oh2023blackvip}: the distinction is determined by parameter accessibility, architectural access, and gradient flow, rather than by adversarial security, privacy protection, or interpretability considerations. Unlike existing settings, Black-CL enforces three primary real-world challenges: (i) \textit{Weight and Architecture Inaccessibility}, which restricts the learner to a minimal set of external components; (ii) \textit{Constrained Computation}, which necessitates high parameter efficiency; and (iii) \textit{Task-Agnostic Inference}, which requires robust discrimination across an expanding global universe of labels. Importantly, Black-CL is stricter than simply freezing a locally accessible backbone. A frozen white-box model may still expose its architecture, intermediate activations, and backward graph, enabling standard gradient-based prompt tuning or the insertion of adapters. In Black-CL, only output embeddings or logits can be queried: gradients cannot propagate through the backbone, and its internal structure cannot be inspected or modified. By formalizing these constraints into a unified evaluation protocol, Black-CL provides a rigorous framework to assess the true deployability of incremental learners in black-box environments. 
In this paper, we propose BETA (\textbf{B}lack-box \textbf{E}mbedding \textbf{T}uning and \textbf{A}daptation), a simple and effective baseline designed for Black-CL. BETA is grounded in the key insight that in the absence of weight access, the complexities of continual learning can be effectively navigated by focusing exclusively on the optimization and adaptation of textual prototypes. This objective is achieved through three synergistic components. First, \textit{Semantic Projection Accumulation} (SPA) incrementally builds a textual knowledge base to capture task-specific features. Second, \textit{Latent Distribution Replay} (LDR) utilizes parametric distribution modeling to anchor the embedding space, thereby preventing catastrophic forgetting with minimal storage overhead. Finally, \textit{Test-Time Prototype Adaptation} (TTPA) introduces an online, instance-aware refinement mechanism that continuously recalibrates decision boundaries over the test stream to mitigate inter-task interference.
Our extensive experiments across ten diverse datasets and various backbones, including CLIP~\cite{radford2021learning}, SigLIP~\cite{zhai2023sigmoid}, and SigLIP 2~\cite{tschannen2025siglip}, demonstrate the superior performance and robustness of BETA. Under both full-shot and 16-shot settings, BETA consistently outperforms existing black-box tuners by significant margins. On the primary full-shot stream, BETA achieves 85.4\% Last Accuracy while preserving strong transfer performance. Remarkably, with only 0.05 M trainable parameters and a 180--3000$\times$ reduction in parameter overhead, BETA remains closely comparable to existing state-of-the-art white-box CL methods.
Our contributions are summarized as follows:
\begin{itemize}
\item We introduce \textbf{Black-CL}, a more realistic benchmark for VLMs that mirrors real-world edge-cloud deployment constraints.
\item We propose \textbf{BETA}, a simple and effective black-box baseline that achieves robust knowledge retention by focusing exclusively on textual prototype optimization.
\item We demonstrate that \textbf{BETA} significantly outperforms existing black-box tuners and remains highly competitive with white-box CL methods.
\end{itemize}
\begin{figure*}[!t]
\centering
    \includegraphics[width=\textwidth]{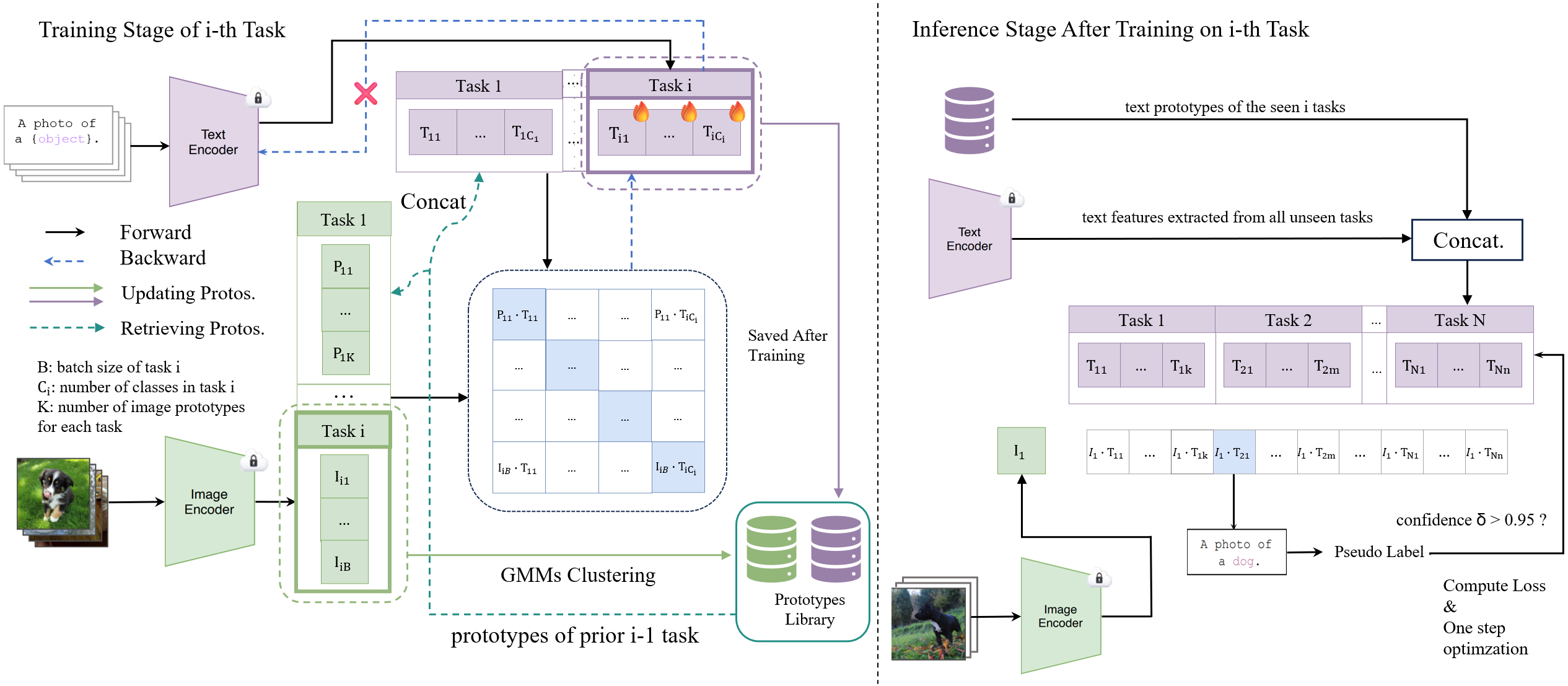}
    \caption{During training stage, BETA incrementally builds a textual prototype library by optimizing class-specific embeddings in the latent
space. To mitigate forgetting without storing raw images, Latent Distribution Replay (LDR) utilizes parametric distribution modeling to anchor the embedding space. At inference stage, BETA concatenate textual prototypes from trained tasks with text features of unseen tasks extracted by a frozen CLIP text encoder, then perform task-id-free inference across all tasks. Additionally, Test-Time Prototype Adaptation (TTPA) introduces an efficient refinement mechanism to dynamically recalibrate decision boundaries.}
    \label{teaser}
\end{figure*}
\section{Related Work}
\noindent\textbf{Continual Learning.}
Continual learning (CL) has expanded its scope from traditional
Task-Incremental Learning (TIL)~\cite{hsu2018re, de2021continual,
wang2024comprehensive, roth2024practitioner} and Class-Incremental Learning
(CIL)~\cite{rebuffi2017icarl, wang2022dualprompt, smith2023coda,
tang2023prompt, kurniawan2024evolving, liu2026ider} to more complex scenarios. In
traditional CIL, existing methods are commonly grouped into memory-based,
regularization-based, and dynamic-based models. Memory-based methods retain
exemplars or replay synthetic/episodic samples to approximate previous data
distributions~\cite{rebuffi2017icarl, lopez2017gradient, buzzega2020dark}.
Regularization-based methods constrain updates to important parameters or
outputs so that new-task learning does not overwrite past
knowledge~\cite{kirkpatrick2017overcoming, zenke2017continual,
aljundi2018memory}. Dynamic-based methods expand, allocate, or mask
task-specific capacity to reduce interference among tasks~\cite{rusu2016progressive,
yoon2018lifelong, mallya2018packnet}. Beyond these conventional settings,
Multi-domain Task-Incremental Learning (MTIL)~\cite{zheng2023preventing}
considers task streams with stronger domain shifts. Recently, Cross-domain Task-Agnostic Incremental Learning (X-TAIL)~\cite{xu2024advancing, luo2025ladascalablelabelspecificclip} takes one step further by discarding task or domain identity, pushing the setting closer to practical scenarios. Under this setting, existing methods often rely on a reference dataset for feature-space knowledge distillation~\cite{zheng2023preventing, yu2024select}, the dependence on incremental model parameter expansion to adapt to new tasks~\cite{tang2024mind, yu2024boosting}, and the utilization of routing mechanism during the inference phase to choose which set of parameters corresponding to a specific task should be used~\cite{xu2024advancing, luo2025ladascalablelabelspecificclip}. Despite the effectiveness of these methods, they fundamentally rely on white-box assumptions, requiring full access to model weights for structural adaptation and knowledge distillation. However, in practical deployment, Large VLMs are increasingly served as proprietary, cloud services or hosted on edge devices where gradient computation and extensive parameter storage are strictly prohibited due to hardware constraints. This significant gap between academic assumptions and industrial reality underscores the importance of Black-box Continual Learning.

\par\noindent\textbf{Black-box Optimization.}
Black-box optimization broadly studies how to optimize an objective when model
internals, parameters, or gradients are inaccessible. Two major application
lines are black-box adversarial attack and black-box learning/tuning. In
score-based black-box adversarial attacks, the attacker has no access to the
target model's internal parameters or gradients and must rely on queried output
scores~\cite{ilyas2018prior, ilyas2018black, huang2019black,
andriushchenko2020square, cheng2019improving}. Such methods commonly employ
zeroth-order or query-efficient search techniques, including natural evolution
strategies (NES)~\cite{wierstra2014natural}, to find input perturbations that
increase the loss and induce misclassification.
Policy gradient~\cite{sutton1999policy}, another black-box optimization
paradigm widely used in reinforcement learning, directly optimizes policies
from sampled returns. Unlike NES, which is typically applied to continuous
search spaces, policy-gradient methods naturally support discrete action
choices and therefore provide a principled way to search over discrete prompts.
\par\noindent Black-box tuning has become an important model-adaptation setting when model weights and gradients are inaccessible. In language models, Black-Box Tuning (BBT)~\cite{sun2022black} and BBTv2~\cite{sun2022bbtv2} optimize continuous prompts via zeroth-order optimization, and BDPL~\cite{diao2022black} explores discrete prompts through reinforcement learning. For vision-language models, BlackVIP~\cite{oh2023blackvip} learns visual prompts embedded in images with zeroth-order updates, while LFA~\cite{ouali2023black} and collaborative black-box tuning~\cite{guo2023black} assume access to backbone outputs or precomputed features to improve alignment or adapt output visual features. Most related to our setting, LCS~\cite{kuwana2024black} studies selective forgetting for black-box PTMs by optimizing a learnable text prompt with derivative-free optimization and proposes Latent Context Sharing to improve the efficiency of prompt parametrization. However, its goal is to \emph{erase} specified classes while preserving the rest, rather than continually \emph{acquire} new cross-domain tasks. Additionally, RAIL~\cite{xu2024advancing} is uniquely positioned as a dedicated continual learning method that incorporates black-box adaptation capabilities. Our use of the term ``black-box'' follows this learning/tuning line: it refers to restricted access to model parameters, architecture, and gradients, rather than the security-oriented setting of black-box adversarial attacks. Accordingly, our objective is not to degrade a model through adversarial inputs, but to optimize external representations that improve classification accuracy under continual learning.
\section{Black-Box Continual Learning for VLMs}
Despite significant strides in continual learning (CL), particularly in class-incremental settings, the majority of existing methods are still evaluated on single and small-scale dataset e.g., CIFAR~\cite{krizhevsky2009learning}. While these approaches demonstrate consistent improvements within controlled environments, their efficacy under restricted model-access settings remains largely underexplored. A growing literature has studied black-box learning and tuning, including BBT~\cite{sun2022black}, BBTv2~\cite{sun2022bbtv2}, BDPL~\cite{diao2022black}, BlackVIP~\cite{oh2023blackvip}, LFA~\cite{ouali2023black}, CBBT~\cite{guo2023black}, and LCS~\cite{kuwana2024black}; however, how to satisfy black-box constraints in continual learning for VLMs remains unexplored. In such settings, a pre-trained VLM may expose only output embeddings or logits, while continual adaptation must be achieved through lightweight external parameters rather than backbone modification.
To bridge this gap, we propose a new benchmark: Black-Box Continual Learning (Black-CL) for VLMs. This benchmark presents three primary challenges:
(i) Weight and Architecture Inaccessibility: The learner has no access to the backbone parameters, internal structure, intermediate activations, or backward graph. Thus, no gradient is allowed to flow through the feature extractor, and the backbone cannot be modified by inserting adapters or other internal modules. This is stricter than freezing an otherwise accessible model and requires adaptation through external components learned solely from queried outputs.
(ii) Constrained Computation: The limited trainable-parameter and optimization budget necessitates high parameter efficiency and low update costs. Methods must maintain a small footprint of tuned parameters while keeping optimization lightweight.
(iii) Task-Agnostic Inference: As the label set expands and tasks span diverse domains, logits from related classes can interfere with one another. This interference often leads to mispredictions, even when utilizing a robust pre-trained backbone. In this section, we present the complete training and inference pipeline for Black-CL.
\subsection{Training Stage}
In real-world scenarios, we expect a model to continuously acquire capabilities across multiple scenes. To simulate this, we treat each dataset as one task and construct a continual learning task stream $\mathcal{S}=\{\mathcal{D}_1,\dots,\mathcal{D}_T\}$, where $\mathcal{D}_t=\mathcal{D}_t^{train}\cup\mathcal{D}_t^{test}$ and the corresponding label space is $\mathcal{Y}_t$. Label spaces are disjoint across tasks, i.e., $\mathcal{Y}_t \cap \mathcal{Y}_{t'}=\emptyset$ for $t\neq t'$. Training proceeds sequentially with task boundaries available to define the current stream.
The VLM backbone is treated as a forward-only interface $\Phi=\{E_I, E_T\}$. The learner can only query output image/text features, without access to encoder parameters, layer structure, intermediate activations, or gradients. We denote the normalized image and text embeddings used for cosine logits as $z=E_I(x)$ and $u=E_T(t)$, respectively. When fitting LDR, $r=E_I^{\mathrm{raw}}(x)$ denotes the corresponding raw CLIP image feature before the final L2 normalization, with $z=r/\|r\|_2$. The queried features are detached outputs: optimization may compute gradients for an external parameter set $\theta_t$ (e.g., textual prototypes), but no computational graph or gradient passes through $E_I$ or $E_T$. Consequently, methods that require backpropagation through the encoder, such as standard gradient-based prompt tuning, or architectural intervention, such as inserting adapters into internal layers, are outside the Black-CL protocol.

\noindent\textbf{Datasets.} Following CoOp~\cite{zhou2022conditional}, we instantiate $\mathcal{S}$ with Caltech101~\cite{fei2004learning}, OxfordPets~\cite{parkhi2012cats}, StanfordCars~\cite{krause20133d}, Flowers102~\cite{nilsback2008automated}, Food101~\cite{bossard2014food}, FGVCAircraft~\cite{maji2013fine}, SUN397~\cite{xiao2010sun}, DTD~\cite{cimpoi2014describing}, EuroSAT~\cite{helber2019eurosat}, and UCF101~\cite{soomro2012ucf101}, which yields a cross-domain sequence with substantial distribution shifts. Notably, while we include MNIST~\cite{deng2012mnist} in Section~\ref{sec:main_results} for a fair comparison with existing white-box methods, it is excluded from our primary Black-CL benchmark due to its limited relevance to complex real-world VLM deployments. 
\subsection{Inference Stage}
In practical deployments, a system cannot predict which specific task a user's input belongs to; therefore, it must possess the capability to perform direct inference within the global label space. This task-agnostic requirement is challenging as the model must discern the correct category from the global universe $\mathcal{Y} = \bigcup_{t=1}^T \mathcal{Y}_t$ without relying on a task oracle. After learning task $i$, unlike Primal-RAIL~\cite{xu2024advancing}  uses a frozen CLIP as task identifier, we directly employ a hybrid classifier $g_i(c)$ that combines learned embeddings for seen classes and frozen zero-shot embeddings for unseen classes. The final prediction is: 
\begin{equation}
\hat{y} = \operatorname*{argmax}_{c \in \mathcal{Y}} \cos(E_I(x), g_i(c)).
\end{equation}

\begin{figure}[t]
\centering
\includegraphics[width=0.86\columnwidth,trim=0 4 8 0,clip]{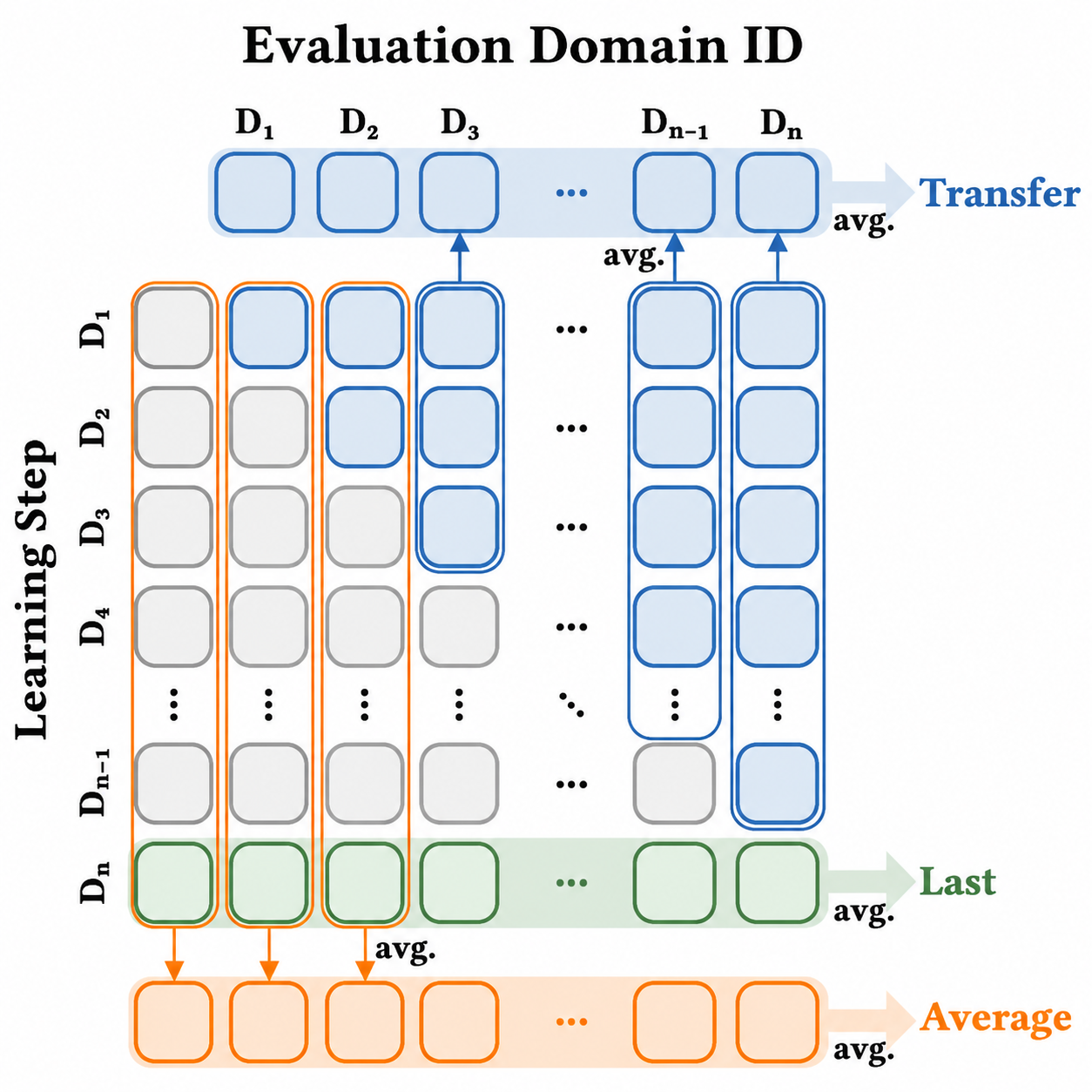}
\caption{\textbf{Matrix visualization of the evaluation metrics.}
The horizontal axis denotes the test task, and the vertical axis denotes the
stage after sequentially training each task in the continual-learning stream.
Each rounded cell represents the accuracy $R_{i,j}$ obtained after learning
task $i$ and testing on task $j$. The upper-triangular future-task cells are
aggregated for Transfer Accuracy; the final-row cells are averaged for Last
Accuracy; and the column-wise summaries indicate averaging over all entries for
Average Accuracy.}
\label{fig:metric_matrix}
\end{figure}

\noindent\textbf{Metrics.} To rigorously evaluate both knowledge retention and
cross-task transfer, we perform a full evaluation on all $T$ test sets after
every training stage. Let $R_{i,j}$ denote the accuracy on task $j$ after the
model has sequentially learned tasks up to task $i$. We report three metrics:
\textit{Transfer Accuracy} measures zero-shot performance on tasks that have
not yet been trained by averaging the future-task entries $R_{i,j}$ with
$j>i$ after each stage and then averaging these stage-wise values over the
stream; \textit{Average Accuracy} measures the mean performance over the whole
evaluation matrix by averaging $R_{i,j}$ over all training stages $i$ and all
test tasks $j$, including both seen and unseen tasks; and \textit{Last
Accuracy} reflects the final performance after the entire stream is completed
by averaging the last-row accuracies $R_{T,j}$ over all test tasks. The
aggregation regions for these metrics are visualized in
Fig.~\ref{fig:metric_matrix}.
\begin{table*}[t]
\centering
\caption{\textbf{Main results on the full-shot black-box continual learning task stream}. We report Transfer Accuracy, Average Accuracy, and Last Accuracy across ten datasets. \textbf{Bold} indicates the best performance among black-box methods. The supplementary material reports the corresponding 16-shot results, where \textbf{BETA still consistently outperforms existing black-box methods.}}
\label{tab:full_shot_results_bcl}
\begin{adjustbox}{width=0.93\textwidth}
\begin{tabular}{l c cccccccccc >{\columncolor{gray!10}}c}
\toprule
& &\multicolumn{11}{c}{Continual Learning Task Stream $\mathcal{S}$} \\
 \cmidrule(lr){3-13}
& Params & \rotatebox{90}{Caltech} & \rotatebox{90}{Pets} & \rotatebox{90}{Cars} & \rotatebox{90}{Flowers} & \rotatebox{90}{Food} & \rotatebox{90}{Aircraft} & \rotatebox{90}{SUN397} & \rotatebox{90}{DTD} & \rotatebox{90}{EuroSAT} & \rotatebox{90}{UCF101} & \cellcolor{white}\rotatebox{90}{\textit{Average}} \\
\midrule
Zero-shot & -- & 72.4 & 88.6 & 65.5 & 70.7 & 85.3 & 24.8 & 59.1 & 33.7 & 45.2 & 64.9 & 61.0 \\ 
\midrule
& &\multicolumn{11}{c}{Transfer Accuracy} \\
 \cmidrule(lr){3-13}
CBBT (Tip-Adapter)~\cite{guo2023black} & 2.2 M & -- & 87.5 & 64.2 & 68.9 & 82.5 & 21.5 & 57.3 & 31.8 & 42.5 & 62.8 & 57.7 \\
LFA~\cite{ouali2023black} & 0.3 M & -- & 87.0 & 63.8 & 67.5 & 81.2 & 20.4 & 55.8 & 30.5 & 41.5 & 61.8 & 56.6 \\
BlackVIP~\cite{oh2023blackvip} & 0.01 M & -- & 85.8 & 62.5 & 66.4 & 79.8 & 19.8 & 54.5 & 29.8 & 40.5 & 60.1 & 55.5 \\
LCS~\cite{kuwana2024black} & 0.002 M & -- & 85.2 & 61.5 & 65.8 & 79.1 & 18.5 & 53.8 & 28.5 & 39.8 & 59.1 & 54.6 \\
Primal-RAIL~\cite{xu2024advancing} & 18.0 M & -- & 88.6 & \textbf{65.5} & \textbf{70.7} & \textbf{85.3} & \textbf{24.8} & 59.1 & 33.7 & 45.2 & 64.9 & 59.7 \\ 
\textbf{BETA (Ours)} & \textbf{0.05 M} & -- & \textbf{88.6} & 62.8 & 69.7 & 76.2 & 23.9 & \textbf{61.9} & \textbf{38.0} & \textbf{49.5} & \textbf{67.2} & \textbf{59.8}  \\
\midrule
& &\multicolumn{11}{c}{Average Accuracy} \\
 \cmidrule(lr){3-13}
CBBT (Tip-Adapter)~\cite{guo2023black} & 2.2 M & 86.2 & 89.5 & 74.5 & 83.2 & 83.8 & 33.1 & 63.5 & 40.2 & 46.5 & 62.1 & 66.3 \\
LFA~\cite{ouali2023black}  & 0.3 M & 84.8 & 88.8 & 73.2 & 81.8 & 82.5 & 31.8 & 61.5 & 38.8 & 45.2 & 61.2 & 65.0 \\
BlackVIP~\cite{oh2023blackvip} & 0.01 M & 82.5 & 87.2 & 71.5 & 80.1 & 81.0 & 30.2 & 59.4 & 37.2 & 43.1 & 59.2 & 63.1 \\
LCS~\cite{kuwana2024black} & 0.002 M & 81.4 & 86.5 & 70.1 & 78.8 & 79.5 & 28.8 & 58.1 & 35.8 & 42.1 & 58.1 & 61.9 \\
Primal-RAIL~\cite{xu2024advancing} & 18.0 M & 83.1 & 92.8 & 79.5 & 87.1 & 86.5 & 35.3 & 65.3 & 44.7 & 49.2 & 63.1 & 68.6  \\
\textbf{BETA (Ours)} & \textbf{0.05 M} & \textbf{92.1} & \textbf{93.3} & \textbf{82.2} & \textbf{89.9} & \textbf{84.5} & \textbf{38.6} & \textbf{67.6} & \textbf{49.3} & \textbf{59.0} & \textbf{68.6} & \textbf{72.5}   \\
\midrule
& &\multicolumn{11}{c}{Last Accuracy} \\
 \cmidrule(lr){3-13}
CBBT (Tip-Adapter)~\cite{guo2023black} & 2.2 M & 83.5 & 82.4 & 70.1 & 85.5 & 81.5 & 29.5 & 59.5 & 36.8 & 52.5 & 70.8 & 65.3 \\
LFA~\cite{ouali2023black} & 0.3 M & 82.8 & 81.5 & 69.2 & 84.8 & 80.8 & 28.8 & 58.5 & 35.5 & 51.5 & 69.8 & 64.3 \\
BlackVIP~\cite{oh2023blackvip} & 0.01 M & 80.5 & 79.8 & 67.2 & 82.5 & 78.5 & 26.8 & 56.5 & 33.5 & 49.5 & 68.1 & 62.3 \\
LCS~\cite{kuwana2024black} & 0.002 M & 79.5 & 78.5 & 65.8 & 81.5 & 77.5 & 25.5 & 55.5 & 32.8 & 48.5 & 66.8 & 61.2 \\
Primal-RAIL~\cite{xu2024advancing} & 18.0 M & 94.3 & 93.5 & 82.1 & 96.2 & 89.0 & 46.1 & 76.4 & 71.4 & 94.2 & 85.2 & 82.8 \\
\textbf{BETA (Ours)} & \textbf{0.05 M} & \textbf{95.5} & \textbf{94.0} & \textbf{87.1} & \textbf{98.8} & \textbf{89.8} & \textbf{53.2} & \textbf{77.2} & \textbf{75.9} & \textbf{95.9} & \textbf{86.4} & \textbf{85.4} \\
\bottomrule
\end{tabular}
\end{adjustbox}
\end{table*}
\begin{table*}[t]
\centering
\caption{\textbf{Comparison with white-box CL methods on the full-shot task stream.} We report Transfer Accuracy, Average Accuracy, and Last Accuracy across ten datasets. Unlike baselines that require full model access ($\times$), \textbf{BETA} operates under strict black-box constraints ($\surd$) and utilizes significantly fewer trainable parameters (only 0.05 M). \textbf{Bold} indicates the best performance among all methods. The supplementary material reports the corresponding 16-shot results, where \textbf{BETA is still competitive with state-of-the-art white-box methods}.}
\label{tab:full_shot_tail}
\begin{adjustbox}{width=0.93\textwidth}
\begin{tabular}{l cc cccccccccc >{\columncolor{gray!10}}c}
\toprule
& & &\multicolumn{11}{c}{Continual Learning Task Stream $\mathcal{S}$} \\
 \cmidrule(lr){4-14}
& Black-Box ? & Params & \rotatebox{90}{Aircraft} & \rotatebox{90}{Caltech} & \rotatebox{90}{DTD} & \rotatebox{90}{EuroSAT} & \rotatebox{90}{Flowers} & \rotatebox{90}{Food} & \rotatebox{90}{MNIST} & \rotatebox{90}{Pets} & \rotatebox{90}{Cars} & \rotatebox{90}{Sun397} & \cellcolor{white}\rotatebox{90}{\textit{Average}} \\
\midrule
Zero-shot & -- & -- & 23.8 & 74.3 &36.4 &37.4 &64.1& 83.4 &43.9 &87.8 &65.5 &60.8 &57.7 \\ 
\midrule
& & &\multicolumn{11}{c}{Transfer Accuracy} \\
 \cmidrule(lr){4-14}
LwF~\cite{li2017learningforgetting}    & $\times$ & 149.6 M & -- & 62.4 & 27.8 & 10.7 & 52.0 & 76.0 & 25.4 & 68.3 & 30.4 & 54.5 & 45.3\\
WiSE-FT~\cite{wortsman2022robust}    & $\times$ & 149.6 M & -- &59.9 & 25.4 & 10.2 &43.9& 67.9& 29.4 & 57.4 &24.1 &50.3 &40.9 \\
ZSCL~\cite{zheng2023preventing}  & $\times$ & 149.6 M & -- & 71.3 & 33.8 & 33.0 & 66.2 & 85.2 & 40.2 &81.9 &57.3 &62.5 &59.0  \\
MoE-Adapters~\cite{yu2024boosting}    & $\times$ & 59.8 M& -- &69.5 & 30.8 & 19.0 & 60.3 & \textbf{85.6} & 43.7 & 85.6 & 55.5 & 57.3 & 56.4  \\
Primal-RAIL~\cite{xu2024advancing}  & $\surd$ &  18.0 M & -- & 74.3 &36.4 &37.4 &64.1& 83.4 &43.9 &87.8 &65.5 &60.8 & 61.5 \\
LADA~\cite{luo2025ladascalablelabelspecificclip} & $\times$ & 9.0 M& -- &75.2 &\textbf{36.1} &36.7& 65.6 &83.9 &\textbf{45.2} &88.0 &\textbf{65.3} &61.1 &61.9 \\
\textbf{BETA (Ours)}    & $\surd$ & \textbf{0.05 M} & -- & \textbf{76.8} & 34.9 & \textbf{47.0} & \textbf{70.8} & 76.4 & 43.1 & \textbf{89.2} & 62.8 & \textbf{65.0} & \textbf{62.9}\\
\midrule
& & &\multicolumn{11}{c}{Average Accuracy} \\
 \cmidrule(lr){4-14}
LwF~\cite{li2017learningforgetting}    & $\times$ & 149.6 M & 25.8 & 81.2 & 48.1 &33.1 &57.0& 75.1 &54.9 &74.5 &35.5 &57.1& 54.2 \\
WiSE-FT~\cite{wortsman2022robust}    & $\times$ & 149.6 M & 14.9 & 79.8 & 45.3 & 17.1 & 55.7 & 70.1 &52.0 &68.0 &35.6& 53.3 & 49.2  \\
ZSCL~\cite{zheng2023preventing}  & $\times$ & 149.6 M & 39.7 & 80.8 & 52.9 &40.8 &79.3 &\textbf{88.0} &51.4 &85.5 &62.9 & 64.1 & 64.5  \\
MoE-Adapters~\cite{yu2024boosting}    & $\times$ & 59.8 M& 52.4 & 79.4 & 57.7 & 42.7 & 81.1 &86.6 &64.8 &86.7 &61.3 &59.0 &67.2 \\
Primal-RAIL~\cite{xu2024advancing} & $\surd$ &   18.0 M & 48.8 & 89.6 & 59.0 & 74.4 & 84.0 & 86.1 & 65.6&  89.5 &68.9 &62.3 &72.8 \\
LADA~\cite{luo2025ladascalablelabelspecificclip} & $\times$ & 9.0 M& \textbf{53.9} & \textbf{93.6}&\textbf{66.6} &78.0 &85.3 &86.7& \textbf{65.2} &89.9 &\textbf{69.7} &62.7 & \textbf{75.2} \\
\textbf{BETA (Ours)}    & $\surd$ & \textbf{0.05 M} & 49.4 & 86.2 & 66.2 & \textbf{81.2} & \textbf{87.4} & 83.0 & 64.3 & \textbf{90.5} & 67.6 & \textbf{66.2}&74.2 \\
\midrule
& & &\multicolumn{11}{c}{Last Accuracy} \\
 \cmidrule(lr){4-14}
LwF~\cite{li2017learningforgetting}    & $\times$ & 149.6 M & 9.6 & 77.1& 55.3& 38.7& 60.5& 83.1 &\textbf{99.5} &85.9& 49.6& 80.0& 63.9  \\
WiSE-FT~\cite{wortsman2022robust}    & $\times$ & 149.6 M & 18.1 &84.9 &53.4 &27.0 &69.6 &88.0& 88.4 &91.5 &76.7 &\textbf{80.2} &67.8  \\
ZSCL~\cite{zheng2023preventing}  & $\times$ & 149.6 M & 33.8 & 80.4 & 60.2 & 31.1 & 85.8 & \textbf{91.3} & 80.4 &93.7 &84.9 &79.0 & 72.1  \\
MoE-Adapters~\cite{yu2024boosting}    & $\times$ & 59.8 M & 51.9 & 79.0 & 64.2 & 51.5 & 95.1 &87.6 &96.4 &89.1 &84.4 &74.0 &77.5 \\
Primal-RAIL~\cite{xu2024advancing}& $\surd$ &    18.0 M & 45.8& 94.1& 70.7& 94.2& 96.5& 89.0& 98.1& 93.5& 82.0 &76.5& 84.0\\
LADA~\cite{luo2025ladascalablelabelspecificclip} & $\times$ & 9.0 M & \textbf{55.5} & \textbf{96.2} & 75.8 & 95.8 &98.4 &89.6 &98.8 &\textbf{94.5} &\textbf{87.3} &77.2 & \textbf{86.9} \\
\textbf{BETA (Ours)}    & $\surd$ & \textbf{0.05 M} & 54.6 & 92.2 & \textbf{75.8} & \textbf{95.9} & \textbf{98.5} & 89.7 & 96.1 & 93.8 & 86.8 & 77.1&86.0 \\
\bottomrule
\end{tabular}
\end{adjustbox}
\end{table*}
\begin{table*}[t]
\centering
\caption{\textbf{Full-shot performance gains of BETA across various backbones.} Each backbone spans multiple rows to accommodate different variants and their corresponding adaptation results. $\Delta$ denotes the gain over the Zero-shot baseline. The supplementary material reports the corresponding 16-shot results, which demonstrate that \textbf{BETA consistently delivers significant accuracy improvements across all datasets upon completing the entire black-box continual learning stream.}}
\label{tab:backbone_comparison_6rows_full}
\begin{adjustbox}{width=0.93\textwidth}
\begin{tabular}{lll cccccccccc c}
\toprule
& & & \multicolumn{11}{c}{\textbf{Continual Learning Task Stream $\mathcal{S}$}} \\
\cmidrule(lr){4-14}
\textbf{Backbone} & \textbf{Variant} & \textbf{Method} & \rotatebox{90}{Caltech} & \rotatebox{90}{Pets} & \rotatebox{90}{Cars} & \rotatebox{90}{Flowers} & \rotatebox{90}{Food} & \rotatebox{90}{Aircraft} & \rotatebox{90}{SUN397} & \rotatebox{90}{DTD} & \rotatebox{90}{EuroSAT} & \rotatebox{90}{UCF101} & \rotatebox{90}{\textit{Average}} \\
\midrule
\multirow{6}{*}{CLIP} & \multirow{3}{*}{B/16-224} 
& Zero-shot & 72.4 & 88.6 & 65.5 & 70.7 & 85.3 & 24.8 & 59.1 & 33.7 & 45.2 & 64.9 & 61.0 \\
& & BETA (Ours) & 95.5 & 94.0 & 87.1 & 98.8 & 89.8 & 53.2 & 77.2 & 75.9 & 95.9 & 86.4 & 85.4 \\
& &  $\Delta$ & \cellcolor{gray!10}\textcolor{gain}{+23.1} & \cellcolor{gray!10}\textcolor{gain}{+5.4} & \cellcolor{gray!10}\textcolor{gain}{+21.6} & \cellcolor{gray!10}\textcolor{gain}{+28.1} & \cellcolor{gray!10}\textcolor{gain}{+4.5} & \cellcolor{gray!10}\textcolor{gain}{+28.4} & \cellcolor{gray!10}\textcolor{gain}{+18.1} & \cellcolor{gray!10}\textcolor{gain}{+42.2} & \cellcolor{gray!10}\textcolor{gain}{+50.7} & \cellcolor{gray!10}\textcolor{gain}{+21.5} & \cellcolor{gray!10}\textcolor{gain}{+24.4} \\
\cmidrule(lr){2-14}
& \multirow{3}{*}{L/14-224} 
& Zero-shot & 75.7 &  93.3 &  76.8 &  79.5 &  90.3 & 32.5 & 64.4 & 41.8 & 61.0 & 73.3 & 68.9 \\
& & BETA (Ours) & 96.8 & 95.5 & 90.9 & 99.2 & 93.1 & 63.2 & 80.8 & 79.4 & 97.2 & 90.0 & 88.6\\
& & $\Delta$ & \cellcolor{gray!10}\textcolor{gain}{+21.1} & \cellcolor{gray!10}\textcolor{gain}{+2.2} & \cellcolor{gray!10}\textcolor{gain}{+14.1} & \cellcolor{gray!10}\textcolor{gain}{+19.7} & \cellcolor{gray!10}\textcolor{gain}{+2.8} & \cellcolor{gray!10}\textcolor{gain}{+30.7} & \cellcolor{gray!10}\textcolor{gain}{+16.4} & \cellcolor{gray!10}\textcolor{gain}{+37.6} & \cellcolor{gray!10}\textcolor{gain}{+36.2} & \cellcolor{gray!10}\textcolor{gain}{+16.7} & \cellcolor{gray!10}\textcolor{gain}{+19.7} \\
\midrule
\multirow{6}{*}{SigLIP} & \multirow{3}{*}{B/16-224} 
& Zero-shot & 85.3 & 92.8 & 89.0 & 81.7 & 88.0 & 29.4 & 65.7 & 55.1 & 31.6 & 72.2 &  69.1 \\
& & BETA (Ours)  & 97.4 & 95.3 & 94.6 & 99.6 & 91.5 & 58.1 & 78.5 & 81.4 & 96.7 & 86.4 & 88.0 \\
& &  $\Delta$ & \cellcolor{gray!10}\textcolor{gain}{+12.1} & \cellcolor{gray!10}\textcolor{gain}{+2.5} & \cellcolor{gray!10}\textcolor{gain}{+5.6} & \cellcolor{gray!10}\textcolor{gain}{+17.9} & \cellcolor{gray!10}\textcolor{gain}{+3.5} & \cellcolor{gray!10}\textcolor{gain}{+28.7} & \cellcolor{gray!10}\textcolor{gain}{+12.8} & \cellcolor{gray!10}\textcolor{gain}{+26.3} & \cellcolor{gray!10}\textcolor{gain}{+65.1} & \cellcolor{gray!10}\textcolor{gain}{+14.2} & \cellcolor{gray!10}\textcolor{gain}{+18.9} \\
\cmidrule(lr){2-14}
& \multirow{3}{*}{So/14-224} 
& Zero-shot & 85.4 & 95.0 & 87.5 & 89.5 & 92.6 & 50.1 & 72.0 & 56.4 & 49.4 & 77.6 & 75.5\\
& & BETA (Ours) & 96.4 & 96.4 & 95.9 & 99.8 & 94.7 & 71.7 & 82.9 & 83.5 & 97.1 & 90.8 & 90.9 \\
& & $\Delta$ & \cellcolor{gray!10}\textcolor{gain}{+11.0} & \cellcolor{gray!10}\textcolor{gain}{+1.4} & \cellcolor{gray!10}\textcolor{gain}{+8.4} & \cellcolor{gray!10}\textcolor{gain}{+10.3} & \cellcolor{gray!10}\textcolor{gain}{+2.1} & \cellcolor{gray!10}\textcolor{gain}{+21.6} & \cellcolor{gray!10}\textcolor{gain}{+10.9} & \cellcolor{gray!10}\textcolor{gain}{+27.1} & \cellcolor{gray!10}\textcolor{gain}{+47.7} & \cellcolor{gray!10}\textcolor{gain}{+13.2} & \cellcolor{gray!10}\textcolor{gain}{+15.4} \\
\midrule
\multirow{6}{*}{SigLIP 2} & \multirow{3}{*}{B/16-224} 
& Zero-shot & 85.8 & 92.7 & 92.5 & 81.6 & 88.8 & 28.0 & 69.4 & 55.1 & 35.8 & 66.2 & 69.6 \\
& & BETA (Ours) & 97.4 & 95.5 & 94.4 & 99.6 & 92.3 & 65.6 & 80.0 & 82.1 & 96.6 & 88.0 & 89.2 \\
& &  $\Delta$ & \cellcolor{gray!10}\textcolor{gain}{+11.6} & \cellcolor{gray!10}\textcolor{gain}{+2.8} & \cellcolor{gray!10}\textcolor{gain}{+1.9} & \cellcolor{gray!10}\textcolor{gain}{+18.0} & \cellcolor{gray!10}\textcolor{gain}{+3.5} & \cellcolor{gray!10}\textcolor{gain}{+37.6} & \cellcolor{gray!10}\textcolor{gain}{+10.6} & \cellcolor{gray!10}\textcolor{gain}{+27.0} & \cellcolor{gray!10}\textcolor{gain}{+60.8} & \cellcolor{gray!10}\textcolor{gain}{+21.8} & \cellcolor{gray!10}\textcolor{gain}{+19.6} \\
\cmidrule(lr){2-14}
& \multirow{3}{*}{So/14-224} 
& Zero-shot & 85.3 & 94.2 & 95.0 & 89.2 & 92.5 & 55.0 & 70.8 & 60.1 & 43.3 & 76.0 & 76.1 \\
& & BETA (Ours) & 97.2 & 96.4 & 96.3 & 99.9 & 95.2 & 79.6 & 83.3 & 84.4 & 96.4 & 92.7 & 92.1 \\
& & $\Delta$ & \cellcolor{gray!10}\textcolor{gain}{+11.9} & \cellcolor{gray!10}\textcolor{gain}{+2.2} & \cellcolor{gray!10}\textcolor{gain}{+1.3} & \cellcolor{gray!10}\textcolor{gain}{+10.7} & \cellcolor{gray!10}\textcolor{gain}{+2.7} & \cellcolor{gray!10}\textcolor{gain}{+24.6} & \cellcolor{gray!10}\textcolor{gain}{+12.5} & \cellcolor{gray!10}\textcolor{gain}{+24.3} & \cellcolor{gray!10}\textcolor{gain}{+53.1} & \cellcolor{gray!10}\textcolor{gain}{+16.7} & \cellcolor{gray!10}\textcolor{gain}{+16.0} \\
\bottomrule
\end{tabular}
\end{adjustbox}
\end{table*}
\section{BETA: Black-box Embedding Tuning and Adaptation}
Before detailing the components of our approach, we clarify that the primary contribution of BETA is not the introduction of complex new architectures, but rather the discovery of a highly efficient and effective baseline for Black-CL. BETA serves as a simple baseline grounded in a key insight: by tuning and adapting only the textual prototypes, VLMs can effectively navigate the challenges of continual learning. In this section, we introduce three components of BETA: Semantic Projection Accumulation, Latent Distribution Replay, and Test-Time Prototype Adaptation.
\subsection{Semantic Projection Accumulation (SPA)}
The core of our approach lies in the incremental construction of a textual knowledge base without modifying or differentiating through the VLM backbone. For each task $i$, we define a set of learnable textual prototypes $\mathbf{P}_i = \{p_{i,k}\}_{k=1}^{|\mathcal{Y}_i|}$, where each $p_{i,k}$ is initialized by querying the text encoder for the class-label embedding $u_k = E_T(t_k)$. During task $i$, image embeddings $z = E_I(x)$ are likewise obtained as detached outputs, and only the external vectors in $\mathbf{P}_i$ receive gradients when minimizing the cross-entropy loss. Neither encoder participates in the backward pass. Once training is complete, $\mathbf{P}_i$ is stored as a persistent semantic anchor. Notably, storing these textual prototypes is extremely lightweight, as each prototype is merely a low-dimensional vector compared to the millions of parameters in the VLM backbone, requiring negligible storage overhead even as the task stream grows. The accumulation aspect ensures that as the stream progresses, the global classifier grows by concatenating these optimized prototypes with the frozen zero-shot embeddings of unseen tasks. This allows the model to accumulate specialized knowledge task-by-task while retaining the broad generalization of the original VLM.
\subsection{Latent Distribution Replay (LDR)} 
\begin{figure*}[t]
\centering
\includegraphics[width=0.98\textwidth]{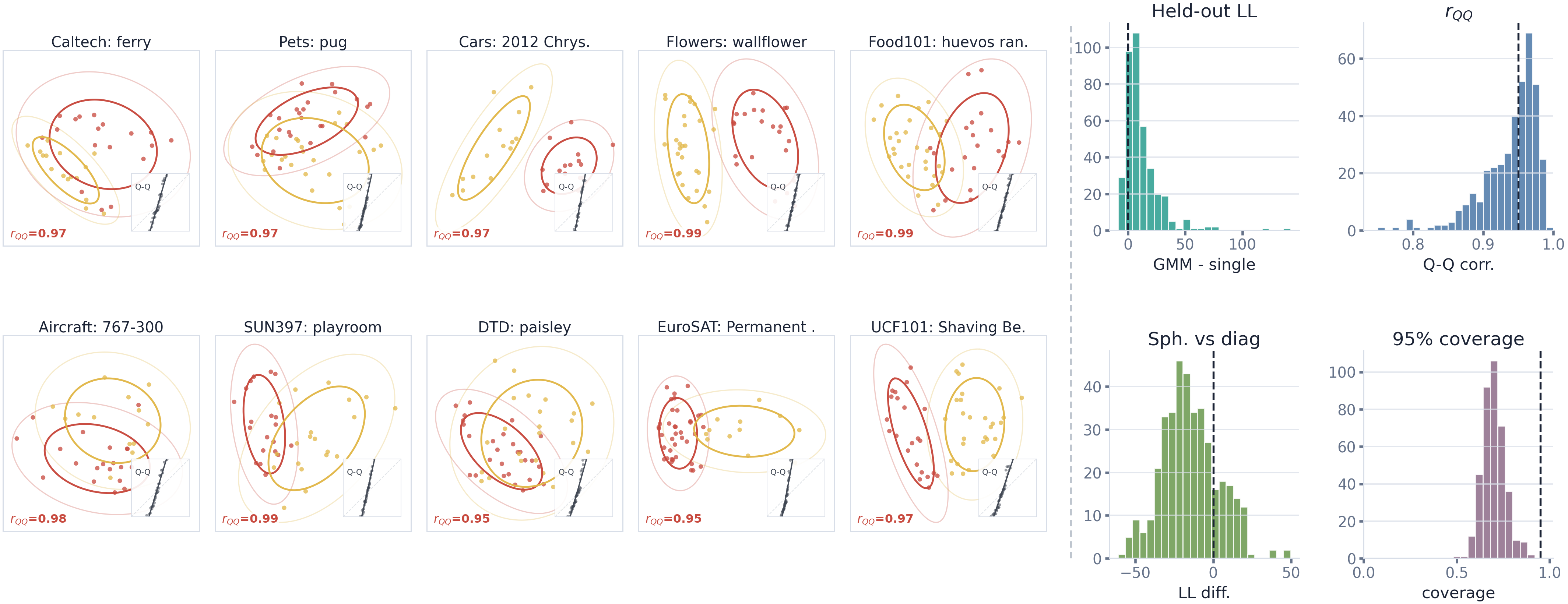}
\caption{\textbf{Visualization and quantitative validation of spherical GMM fitting on CLIP image features.}
Left: for each dataset, we select one representative class and fit a spherical
GMM to its frozen CLIP image features in the original feature space. Points
show the first two PCA dimensions, and colors denote component assignments.
Ellipses visualize the empirical 2D spread of samples assigned to each
spherical component. The inset shows the full-dimensional spherical radial Q-Q
diagnostic; the dashed line denotes the ideal calibrated reference, while the
solid line shows the fitted Q-Q trend. Overall, the left panels suggest that
intra-class CLIP features often exhibit multiple compact local modes, and that
these modes show strong radial Q-Q trends under the spherical approximation.
Right: quantitative diagnostics over all evaluated classes. Held-out LL is
plotted as the histogram of held-out log-likelihood differences between a
spherical GMM and a single spherical Gaussian. \(r_{QQ}\) is plotted as the
histogram of full-dimensional spherical radial Q-Q correlations across classes.
Sph. vs diag is plotted as the histogram of held-out log-likelihood differences
between spherical and diagonal GMMs. 95\% coverage is plotted as the histogram
of empirical sample fractions falling within the theoretical 95\% spherical
chi-square radius. The imperfect 95\% coverage indicates that CLIP features are
not exactly calibrated spherical Gaussians, while the high Q-Q correlations and
the held-out likelihood gains over a single spherical Gaussian support
spherical GMM as a compact local approximation.}
\label{fig:spherical_gmm_validation}
\end{figure*}

To mitigate catastrophic forgetting under black-box constraints, LDR leverages parametric summaries to maintain the discriminability of the global label space. Since the backbone is inaccessible and frozen, the image feature space remains static; however, the expanding label space induces severe inter-task logit confusion. To address this, we capture the statistical essence of each class $c \in \mathcal{Y}_{<i}$ via spherical Gaussian Mixture Models (GMMs). These summaries, denoted as $\Theta_c = \{\pi_{c,k}, \mu_{c,k}, \sigma^2_{c,k}\}_{k=1}^K$, where each component has covariance $\sigma^2_{c,k}I$, represent the component weights, mean vectors, and scalar variances of the original visual manifold without storing raw images.
The use of a GMM is motivated by the geometry of visual representations
extracted by the frozen CLIP image encoder. Recent theoretical analysis
suggests that contrastive objectives such as InfoNCE can encourage
approximately Gaussian feature distributions under certain assumptions
~\cite{betser2026infonce}; related work also studies Gaussian feature
distributions as a useful geometry for representation learning
~\cite{balestriero2025lejepa}. These results do not imply that CLIP image
features are exactly Gaussian, but they provide motivation for using Gaussian
components as compact local approximations. Because a single Gaussian may be
too restrictive for a visual category, we use $K=4$ components to capture
dominant intra-class modes, such as variations in viewpoint and visible object
parts. This compact mixture provides sufficient flexibility in the frozen
CLIP image-feature space while keeping fitting, storage, and replay costs
small. We further validate this spherical GMM approximation empirically in
the analysis in Sec.~\ref{sec:analysis} and Fig.~\ref{fig:spherical_gmm_validation}.
We fit the spherical GMM on the raw outputs $r=E_I^{\mathrm{raw}}(x)$ of the
frozen CLIP image encoder before L2 normalization. Samples drawn from the GMM
are treated as pseudo image features $\tilde{r}$ in this raw feature space and
are L2-normalized as $\tilde{z}=\tilde{r}/\|\tilde{r}\|_2$, together with real
image features, before computing cosine logits with normalized text features.
During the training of task $i$, we obtain pseudo-embeddings $\tilde{z}$ by
sampling $\tilde{r}$ from these stored GMMs and applying the same L2
normalization. Crucially, to ensure global calibration and prevent new prototypes $\mathbf{P}_i$ from encroaching upon the semantic regions of previous categories, we optimize the current prototypes within a unified label space. Unlike the disconnected objective in traditional replay, our joint objective is formulated as:
\begin{equation}
\mathcal{L}_{total} = \mathcal{L}_{CE}(Z_i \cup \tilde{Z}_{1:i-1}, \mathbf{P}_{1:i}),
\end{equation}
where $\mathbf{P}_{1:i}$ denotes the concatenation of both current and previously learned prototypes. By computing cross-entropy over the union of current samples $Z_i$ and replayed samples $\tilde{Z}_{1:i-1}$, the mechanism creates an explicit semantic firewall. This ensures that the adaptation to new concepts does not erode the discriminability of the global label space while maintaining minimal storage overhead.
\subsection{Test-Time Prototype Adaptation (TTPA)}
In task-agnostic inference, the model must discriminate among classes from all tasks within a unified label space $\mathcal{Y}$ without a task oracle. Static decision boundaries often struggle with global miscalibration due to the distribution shift between training and test streams. To address this, we propose TTPA, which performs instance-aware refinement by dynamically recalibrating prototypes based on the local distribution of the test batch.
Let $\mathbf{P}^{(0)}$ denote the prototypes obtained after training. For the
$b$-th incoming test batch $\mathcal{B}_b=\{x_j\}_{j=1}^{N}$, we extract
visual features $\mathbf{f}_j$ and identify reliable anchors
$\mathcal{X}_{a}^{(b)} \subseteq \mathcal{B}_b$ using the current prototypes,
where the subscript $a$ denotes the anchor subset:
\begin{equation}
    \mathcal{X}_{a}^{(b)}
    =
    \left\{
    x_j \in \mathcal{B}_b
    \mid
    \max \operatorname{softmax}\!\left(\mathbf{f}_j(\mathbf{P}^{(b-1)})^\top\right)
    \geq \delta
    \right\},
\end{equation}
where $\delta=0.95$ is the confidence threshold. To satisfy strict inference
latency requirements, TTPA performs at most one gradient step for each batch:
\begin{equation}
    \mathbf{P}^{(b)}
    =
    \begin{cases}
    \mathbf{P}^{(b-1)}
    -\eta\left.\nabla_{\mathbf{P}}\mathcal{L}_{a}^{(b)}
    \right|_{\mathbf{P}=\mathbf{P}^{(b-1)}},
    & |\mathcal{X}_{a}^{(b)}|>0,\\
    \mathbf{P}^{(b-1)},
    & |\mathcal{X}_{a}^{(b)}|=0,
    \end{cases}
\end{equation}
where
$\mathcal{L}_{a}^{(b)}
=\sum_{x_j\in\mathcal{X}_{a}^{(b)}}
\mathcal{L}_{CE}(\operatorname{softmax}(\mathbf{f}_j\mathbf{P}^{\top}),
\tilde{\mathbf{y}}_j)$ and $\tilde{\mathbf{y}}_j$ denotes a smoothed
pseudo-label obtained from the predicted class, with label-smoothing
coefficient $\epsilon=0.1$. Thus, if a batch contains no reliable anchor, the
update is skipped. This is a conservative confidence gate: the model retains its
current prototypes rather than adapting to uncertain pseudo-labels. Following
the two-forward protocol used in the original evaluation, the final logits for
$\mathcal{B}_b$ are then recomputed using the post-update prototypes
$\mathbf{P}^{(b)}$.
Unlike batch-wise transient adaptation, TTPA is \textbf{continuous within a
complete evaluation pass}. The updated state $\mathbf{P}^{(b)}$ initializes
the next batch, and updates are not reset at batch or task-dataset boundaries.
The adapted copy is reset to $\mathbf{P}^{(0)}$ only after all test batches in
the evaluation pass have been processed. Consequently, reliable information
accumulates throughout test-time adaptation, while no test-derived state is
carried into training or a subsequent independent evaluation. The comparison
in Fig.~\ref{fig:ttpa_reset_strategy}(a) validates this continuous protocol.
\section{Experiments}
\providecommand{\tbd}{\textcolor{red}{\textbf{TBD}}}
\providecolor{gain}{RGB}{0,128,0}

\noindent\textbf{Baseline Methods.} 
To verify the effectiveness of BETA, we conduct comparisons across two regimes: (i) We compare BETA against  black-box tuning methods, including CBBT~\cite{guo2023black}, LFA~\cite{ouali2023black}, BlackVIP~\cite{oh2023blackvip}, and LCS~\cite{kuwana2024black}. Since these methods were originally designed for few-shot or domain adaptation rather than continual learning, they lack inherent mechanisms to handle inter-task interference in a global label space. 
(ii) We further compare BETA against state-of-the-art white-box continual learning method to situate its performance within the broader continual learning landscape. Notably, Primal-RAIL is uniquely positioned as a continual learning method with black-box adaptation capabilities; thus, we include it in both Table~\ref{tab:full_shot_results_bcl} and Table~\ref{tab:full_shot_tail} as a primary competitor. Since Dual-RAIL needs to store all features, it is not possible to complete training under the
full-shot setting because of computational overhead and GPU memory constraint. Additionally, we conduct these comparisons under a 16-shot setting to further evaluate performance. Due to space constraints, these extended experimental results are provided in the supplementary material.

\noindent\textbf{Implementation Details.}
We use CLIP ViT-B/16 as the frozen VLM backbone with input resolution $224$ and perform all learning in the embedding space. For each task, we optimize only the textual prototypes using AdamW (weight decay $5\times 10^{-4}$) with batch size $64$, learning rate $5\times 10^{-3}$ and $200$ epochs for default.
For LDR, we fit a spherical Gaussian mixture model per class with $K=4$ components and weight replayed samples with a default coefficient of $64$. For TTPA, we use confidence filtering with threshold $\delta=0.95$, learning rate $\eta=5e-4$, label smoothing $\epsilon=0.1$, and SGD with momentum $0.9$. At most one prototype update is applied per batch when reliable anchors are available. More details are provided in the supplementary material.
\subsection{Main Results}
\label{sec:main_results}

\noindent\textbf{Comparisons with Black-box Tuning Methods.} As illustrated in Table~\ref{tab:full_shot_results_bcl}, BETA significantly outperforms existing black-box tuning methods across all metrics. Regarding \textit{Last Accuracy}, which reflects resistance to forgetting, BETA achieves 85.4\%. This result surpasses specialized tuners like CBBT (65.3\%) and BlackVIP (62.3\%), highlighting the vulnerability of standard methods to catastrophic forgetting. Compared to Primal-RAIL~\cite{xu2024advancing}, BETA leads in average accuracy by 3.9\% while requiring 0.05 M trainable parameters. The supplementary material provides detailed trainable-parameter calculations for BETA and the baselines. This represents a 360$\times$ reduction in memory overhead. Furthermore, BETA's \textit{Transfer Accuracy} remains close to the zero-shot baseline, demonstrating its ability to enhance task-specific performance while preserving the cross-domain generalization of the original VLM. Notably, consistent performance gains are also observed in the 16-shot setting reported in the supplementary material, which validates the robustness of BETA across varying data scales. We additionally repeat the full-shot experiment using the alternative dataset order adopted by LADA~\cite{luo2025ladascalablelabelspecificclip} and Primal-RAIL~\cite{xu2024advancing} with all hyperparameters fixed; BETA continues to deliver compelling performance across all three reported metrics, as detailed in the supplementary material.

\noindent\textbf{Comparisons with White-box CL Methods.} Table~\ref{tab:full_shot_tail} and the lower-left panel of Fig.~\ref{fig:sliding_window} compare BETA against white-box CL methods that require full model access. Despite operating under strict black-box constraints, BETA achieves highly competitive performance, particularly in terms of \textit{Last Accuracy} (86.0\%) and \textit{Average Accuracy} (74.2\%). These results are on par with or even exceed those of advanced white-box architectures like LADA~\cite{luo2025ladascalablelabelspecificclip} (86.9\%) and MoE-Adapters~\cite{yu2024boosting} (77.5\%), while utilizing significantly fewer trainable parameters. Specifically, BETA requires only 0.05 M parameters, representing a 180$\times$ to 3000$\times$ reduction in parameter scale compared to LADA (9.0 M) and ZSCL (149.6 M). Furthermore, BETA maintains a superior \textit{Transfer Accuracy} (62.9\%), outperforming all white-box baselines. This indicates that while white-box methods often sacrifice the model's inherent zero-shot capabilities through intensive weight updates, BETA effectively preserves and even enhances the VLM's generalization via lightweight prototype tuning and adaptation. Similar competitive trends are observed in the supplementary 16-shot results, further establishing BETA as a robust and efficient alternative to traditional white-box CL approaches.

\noindent\textbf{Impact of Different Backbones.} To verify the generalizability of BETA, we evaluate its performance across various state-of-the-art VLM backbones, including CLIP~\cite{radford2021learning}, SigLIP~\cite{zhai2023sigmoid}, and the latest SigLIP 2~\cite{tschannen2025siglip}. As demonstrated in Table~\ref{tab:backbone_comparison_6rows_full} and the bottom-right panel of Fig.~\ref{fig:sliding_window}, BETA consistently achieves substantial accuracy gains over the zero-shot baseline across all tested architectures and variants. Specifically, using the CLIP-B/16 backbone, BETA improves the average accuracy from 61.0\% to 85.4\%, a remarkable absolute gain of {24.4\%}. Even with more powerful backbones like SigLIP 2-So/14, which already possesses a strong zero-shot foundation (76.1\%), BETA further elevates the performance to 92.1\%. Notably, the gains are particularly pronounced in specialized domains; for instance, on the EuroSAT~\cite{helber2019eurosat} dataset, BETA consistently provides gains exceeding 50\% across multiple backbones. These results indicate that BETA's effectiveness is not tied to a specific architecture but rather stems from its robust adaptation mechanism, making it a versatile solution for enhancing diverse pre-trained VLMs in continual learning scenarios. Similar consistent improvements are documented in the supplementary 16-shot results.
\begin{table}[t]
\centering
\caption{\textbf{Impact of Hyper-parameters on TTPA.} Bold indicates the default setting used in our experiments.}
\label{tab:hyperparams}
\begin{adjustbox}{width=0.4\textwidth}
\begin{tabular}{l ccc}
\toprule
\textbf{Hyper-parameter} & \textbf{Value} & \textbf{Avg. Acc} & \textbf{Last. Acc}  \\
\midrule
\multirow{4}{*}{TTA lr $\eta$} & 0.0001 & 70.7  & 81.4  \\
                                  & \textbf{0.0005} & \textbf{70.8} & \textbf{81.5}  \\
                                  & 0.001  & 69.6  & 81.6  \\
                                  & 0.005   & 62.5  & 81.7  \\
\midrule
\multirow{4}{*}{Confidence threshold $\delta$} & 0.8 & 68.3 & 81.4  \\
                                    & 0.85 & 68.7  & 81.4  \\
                                    & 0.9 & 70.2 & 81.5 \\
                                     & \textbf{0.95} & \textbf{70.8} & \textbf{81.5}  \\
\bottomrule
\end{tabular}
\end{adjustbox}
\end{table}
\begin{figure}[t]
\centering
\includegraphics[width=0.98\columnwidth]{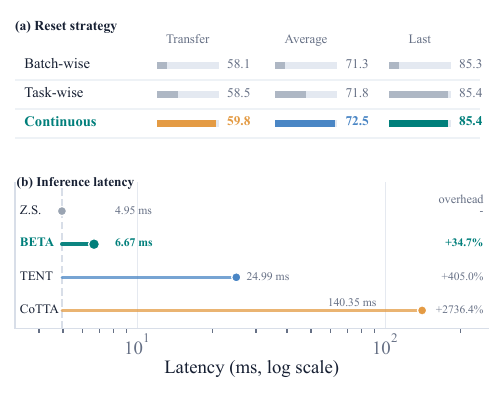}
\caption{\textbf{Compact analysis of TTPA retention and latency.}
(a) Transfer, Average, and Last Accuracy under different prototype-retention
strategies on the full-shot Black-CL stream. Continuous adaptation achieves
the best Transfer and Average Accuracy while matching the best Last Accuracy.
(b) Inference latency on a logarithmic scale, with overhead measured relative
to the zero-shot baseline. BETA remains much closer to zero-shot inference
than TENT~\cite{wang2020tent} and CoTTA~\cite{wang2022continual}.}
\label{fig:ttpa_reset_strategy}
\label{fig:tta_latency}
\end{figure}
\subsection{Analysis}
\label{sec:analysis}
\begin{table}[t]
\centering
\caption{\textbf{Ablation study of BETA components.} We evaluate the contribution of the SPA, LDR and TTPA.}
\label{tab:ablation}
\begin{adjustbox}{width=0.4\textwidth}
\begin{tabular}{l ccc cc}
\toprule
\textbf{Settings} & \textbf{SPA} & \textbf{LDR} & \textbf{TTPA} & \textbf{Avg. Acc} & \textbf{Last. Acc} \\
\midrule
\multirow{4}{*}{16-shot}
& $\surd$  & $\times$ & $\times$ & 68.0  & 77.1 \\
& $\surd$  & $\surd$  & $\times$ & 70.1  & 81.4 \\
& $\surd$  & $\surd$  & $\surd$  & 70.8  & 81.5 \\
\midrule
\multirow{4}{*}{Full-shot}
& $\surd$  & $\times$ & $\times$ & 68.2  & 78.4 \\
& $\surd$  & $\surd$  & $\times$ & 71.7  & 85.3 \\
& $\surd$  & $\surd$  & $\surd$  & 72.5  & 85.4 \\
\bottomrule
\end{tabular}
\end{adjustbox}
\end{table}

\noindent\textbf{Ablations.} We conduct an ablation study to investigate the contribution of each core component in BETA: SPA, LDR, and TTPA. As shown in Table~\ref{tab:ablation}, each module consistently improves performance under both 16-shot and full-shot settings. Starting from the zero-shot baseline, the inclusion of SPA provides initial gains by specializing textual prototypes for new tasks. LDR yields the most significant performance leap, particularly in \textit{Last Accuracy}, which increases by 6.9\% in the full-shot setting. This highlights the critical role of latent distribution replay in stabilizing the embedding space and mitigating catastrophic forgetting. Finally, TTPA further refines the results by adapting prototypes to the test-time distribution, leading to the best overall performance. The consistent trend across different data scales demonstrates that the three components work synergistically to balance knowledge acquisition, retention, and localized adaptation.

\noindent\textbf{Effectiveness of Spherical GMM Fitting.}
To justify the use of spherical GMMs, we analyze the geometry of frozen CLIP
image features at the class level. For each class, we fit a separate spherical
GMM in the original CLIP feature space. The projected visualization shows that
samples assigned to the same component form compact local regions, suggesting
that intra-class variations can be decomposed into several local modes. We
further examine the full-dimensional radial distances within each component
using Q-Q diagnostics. Although the features are not expected to follow an
exactly calibrated spherical Gaussian distribution, the high radial Q-Q
correlation indicates that spherical components provide a reasonable local
approximation to the radial structure of CLIP features. Quantitatively,
spherical GMM improves held-out log-likelihood over a single spherical Gaussian
in 85.7\% of evaluated classes, with an average gain of 11.75 and a median gain
of 7.46. BIC also favors the spherical GMM in 73.8\% of classes, indicating
that the mixture structure is not merely due to over-parameterization. Although
diagonal GMM may achieve better density fit, its additional covariance degrees
of freedom do not translate into better replay performance under the same
downstream protocol and even lead to slight accuracy degradation. Under the
same full-shot CLIP-B/16 setting, spherical GMM achieves 59.7/72.5/85.4
Transfer/Average/Last accuracy, compared with 58.7/72.5/85.3 for diagonal GMM.
The same trend holds in the 16-shot setting, where spherical GMM achieves
59.8/70.8/81.5 compared with 59.6/70.7/81.4 for diagonal GMM. Therefore,
spherical GMM is chosen for the accuracy-storage-stability trade-off, and these
results support it as a compact and effective approximation for modeling
class-wise CLIP image features.
In summary, spherical GMM is not used because CLIP features are exactly
spherical Gaussian, but because it provides a compact, stable, and empirically
effective local approximation to class-wise feature geometry.

\noindent\textbf{Computational Overhead of LDR.}
The Latent Distribution Replay (LDR) module utilizes spherical Gaussian Mixture Models (GMMs) with $K=4$ components per class. As discussed in the method, this choice is well matched to the approximately Gaussian geometry encouraged by contrastive representation learning while retaining enough modes to describe major intra-class variations. We quantify its overhead as follows: (i) Fitting Cost: fitting GMMs via the Expectation-Maximization (EM) algorithm is performed only once at the end of each task. For a dataset with 100 classes, this process takes less than 5 seconds on a standard CPU, making it negligible compared to the prototype optimization phase. (ii) Storage Scalability: For each class, we store $K \times (D + 2)$ scalar values, corresponding to a $D$-dimensional mean vector, one spherical variance, and one mixture weight for each component. For 1,000 classes with $D=1,024$, this amounts to $1,000 \times 4 \times (1,024 + 2) \times 4 \text{ bytes} \approx \mathbf{16.4 \text{ MB}}$. LDR therefore provides a structured, fixed-size replay representation that avoids maintaining a raw-image replay buffer and scales gracefully to thousands of classes. \\
\begin{table}[t]
\centering
\caption{\textbf{GMM overfitting analysis and replay controls in the 16-shot
setting.} The upper part reports per-class GMM log-likelihoods when 16 samples
are used for fitting and the remaining samples for validation. The lower part
reports 16-shot Black-CL performance under different replay constructions;
drops relative to the default $K=4$ GMM replay are shown in parentheses.}
\label{tab:ldr_overfit_noise}
\footnotesize
\setlength{\tabcolsep}{2.6pt}
\begin{tabular*}{\columnwidth}{@{\extracolsep{\fill}}lcccc@{}}
\toprule
\textbf{Dataset} & \textbf{$K=4$ Tr.} & \textbf{$K=4$ Val.} &
\textbf{$K=2$ Tr.} & \textbf{$K=2$ Val.} \\
\midrule
Caltech & 371.7 & -1.8 & 132.0 & 40.5 \\
Pets & 353.3 & 31.3 & 150.6 & 52.0 \\
Flowers & 462.1 & 171.0 & 288.2 & 199.4 \\
\midrule
\textbf{Replay} & \textbf{Trans.} & \textbf{Avg.} &
\textbf{Last} & \textbf{Form} \\
\midrule
\textbf{$K=4$} & \textbf{59.8} & \textbf{70.8} & \textbf{81.5} &
$1{\times}4$ comps. \\
Two $K=2$ & 59.7 \textbf{(-0.1)} & 70.8 & 81.5 & $2{\times}2$ comps. \\
Noise & 59.5 \textbf{(-0.3)} & 69.9 \textbf{(-0.9)} &
80.0 \textbf{(-1.5)} &
mean/var. \\
\bottomrule
\end{tabular*}
\end{table}

\noindent\textbf{GMM Overfitting in 16-Shot LDR.}
We further investigate whether the $K=4$ GMM used by LDR suffers from
small-sample overfitting when only 16 training samples are available per class.
For each class, we fit GMMs on the 16-shot training features and use the
remaining samples for validation. As shown in Table~\ref{tab:ldr_overfit_noise},
$K=4$ obtains much higher training log-likelihood but poorer validation
log-likelihood than $K=2$ on representative datasets, and validation AIC/BIC
scores, standard information criteria for balancing data fit and model
complexity~\cite{stoica2004model}, show the same trend. This confirms that
$K=4$ fits the limited training
features too tightly and generalizes worse to held-out features. However, this
statistical overfitting does not imply worse continual learning performance.
When replacing the default $K=4$ GMM with a budget-matched variant that fits two
separate $K=2$ GMMs per class, Average and Last Accuracy remain unchanged while
Transfer Accuracy drops by 0.1 point. Therefore, the less overfitted GMM fit is
not necessarily the best replay construction for preserving decision
boundaries.
We also test whether LDR's gains merely come from random-noise regularization,
which has been shown to benefit transfer learning in related settings
~\cite{zhong2020regularizing}. Specifically, we replace GMM-sampled
pseudo-features with Gaussian noise that matches the same class-wise mean and
variance. This control still provides mild replay, but its performance drops
substantially compared with GMM replay, especially on Last Accuracy. These
results indicate that LDR relies on the structural information captured by the
mixture components rather than acting only as generic noise regularization. \\

\noindent\textbf{Further Analysis of TTPA.} 
We first analyze the sensitivity of TTPA to two key hyper-parameters: the test-time learning rate $\eta$ and the confidence threshold $\delta$. As shown in Table~\ref{tab:hyperparams}, BETA remains robust within a reasonable parameter range. For the learning rate $\eta$, a value of 0.0005 achieves the optimal balance. Regarding $\delta$, a high threshold (e.g., 0.95) is essential to filter out noisy pseudo-labels, thereby preventing confirmation bias during test-time adaptation. 
We further compare how long the adapted prototypes are retained. As shown in
Fig.~\ref{fig:ttpa_reset_strategy}(a), resetting after every batch discards useful
information and yields 58.1\%, 71.3\%, and 85.3\% in Transfer, Average, and
Last Accuracy, respectively. Retaining updates within each test dataset
provides a modest improvement, whereas continuous adaptation throughout the
complete evaluation pass achieves the best overall results (59.8\%, 72.5\%,
and 85.4\%). These results support accumulating reliable test-time updates
rather than treating each batch independently. Batches without reliable
anchors still leave the prototype state unchanged, preventing uncertain
samples from causing parameter drift.
Furthermore, we evaluate the inference latency of TTPA to quantify its computational cost during test-time adaptation. As reported in Fig.~\ref{fig:tta_latency}(b), compared to existing classic TTA methods like TENT~\cite{wang2020tent} and CoTTA~\cite{wang2022continual}, BETA is significantly more efficient. While CoTTA incurs an overwhelming 2736.4\% overhead, BETA only adds 34.7\% additional latency relative to the zero-shot baseline. This efficiency stems from BETA's lightweight one-step prototype update, which avoids the heavy gradient computations across backbones required by white-box TTA methods and makes the method suitable for resource-limited continual adaptation. Detailed analysis and implementation information are provided in the supplementary material.
\section{Conclusion}
In this paper, we introduce Black-CL, a more practically relevant and challenging benchmark for continual learning in VLMs. By addressing the core constraints of inaccessible model weights, limited compute, and task-agnostic inference, Black-CL narrows the gap between white-box continual learning assumptions and restricted-access model settings. To address these challenges, we propose BETA, a simple and effective baseline. Our findings reveal that by focusing solely on the tuning and adaptation of textual prototypes, BETA achieves performance comparable to state-of-the-art white-box methods. With its minimal overhead and low inference latency, BETA serves as a robust and efficient alternative to traditional continual learning approaches, paving the way for scalable continual adaptation of foundation models under black-box constraints.
\section*{Conflicts of Interest}
The authors declare that they have no conflicts of interest.
\section*{Data and Code Availability}
All datasets used in this study are publicly available from the original
sources cited in the paper. The code will be made available soon.
\bibliographystyle{IEEEtran}
\bibliography{references}
\newcommand{\placeholderbio}[2]{
  Biography placeholder. Add the author's degree information, current
  position, research interests, professional memberships, and selected honors
  here. Photo placeholder: \texttt{#2}.
}
\newcommand{\authorbiography}[2]{
  \IfFileExists{#2}{
    \begin{IEEEbiography}[{\includegraphics[width=1in,height=1.25in,clip,keepaspectratio]{#2}}]{#1}
    \placeholderbio{#1}{#2}
    \end{IEEEbiography}
  }{
    \begin{IEEEbiographynophoto}{#1}
    \placeholderbio{#1}{#2}
    \end{IEEEbiographynophoto}
  }
}
\newcommand{\weihangfangbiography}{
  \IfFileExists{weihang-fang.jpg}{
    \begin{IEEEbiography}[{\includegraphics[width=1in,height=1.25in,clip,keepaspectratio]{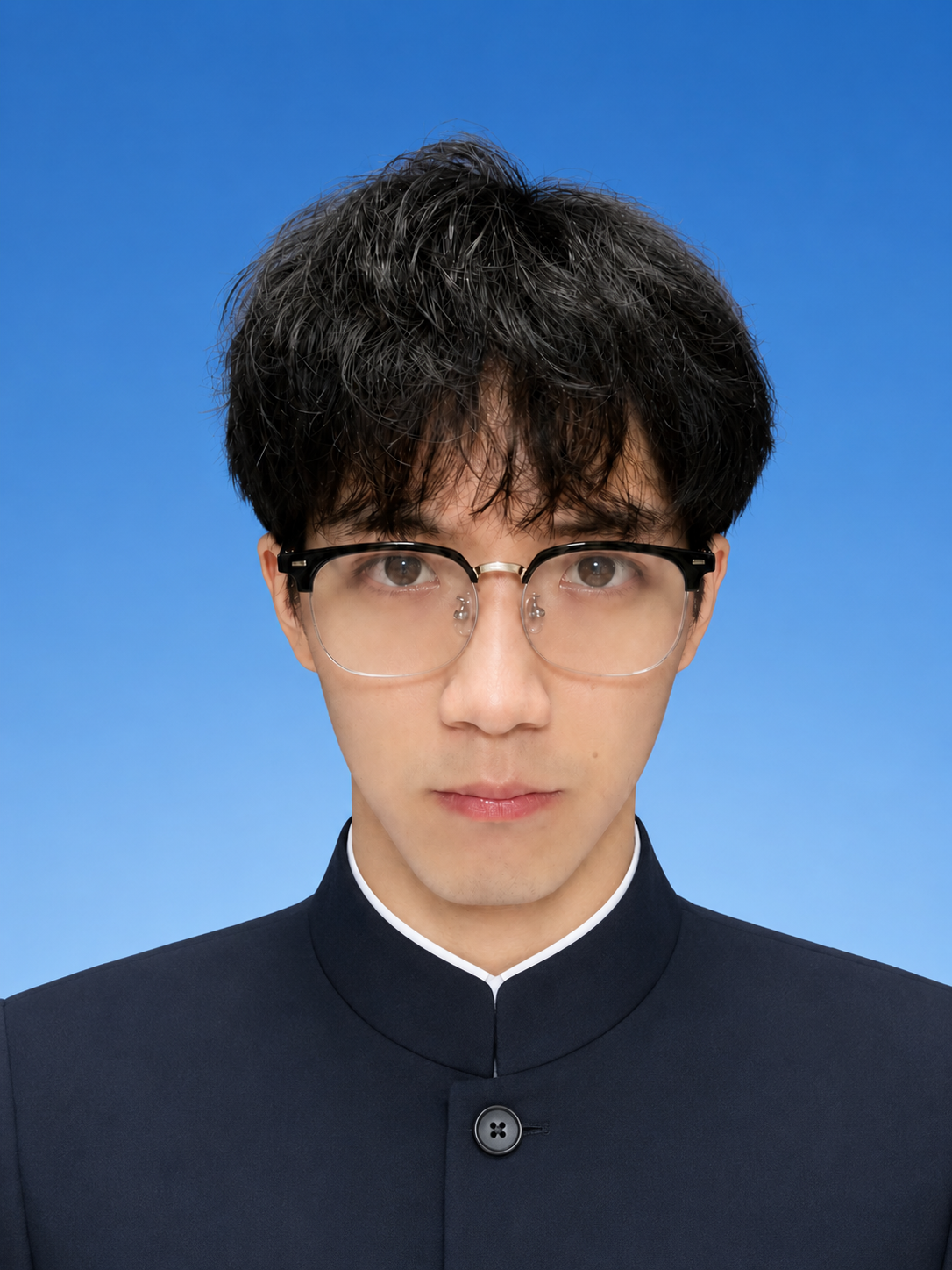}}]{Weihang Fang}
    is currently a junior undergraduate student in the School of Automation
    and Intelligent Sensing, Shanghai Jiao Tong University, Shanghai, China.
    His research interests include continual learning for vision-language
    models.
    \end{IEEEbiography}
  }{
    \begin{IEEEbiographynophoto}{Weihang Fang}
    is currently a junior undergraduate student in the School of Automation
    and Intelligent Sensing, Shanghai Jiao Tong University, Shanghai, China.
    His research interests include continual learning for vision-language
    models.
    \end{IEEEbiographynophoto}
  }
}
\newcommand{\haoyuangaobiography}{
  \IfFileExists{haoyuan-gao.jpg}{
    \begin{IEEEbiography}[{\includegraphics[width=1in,height=1.25in,clip,keepaspectratio]{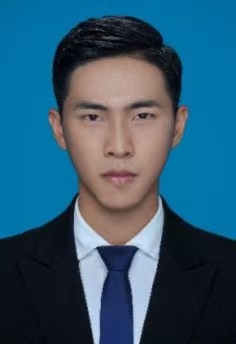}}]{Haoyuan Gao}
    is currently a Ph.D. candidate in the School of Computer Science,
    Shanghai Jiao Tong University, Shanghai, China. His research interests
    include continual learning for vision-language models.
    \end{IEEEbiography}
  }{
    \begin{IEEEbiographynophoto}{Haoyuan Gao}
    is currently a Ph.D. candidate in the School of Computer Science,
    Shanghai Jiao Tong University, Shanghai, China. His research interests
    include continual learning for vision-language models.
    \end{IEEEbiographynophoto}
  }
}
\newcommand{\linghekongbiography}{
  \IfFileExists{linghe-kong.png}{
    \begin{IEEEbiography}[{\includegraphics[width=1in,height=1.25in,clip,keepaspectratio]{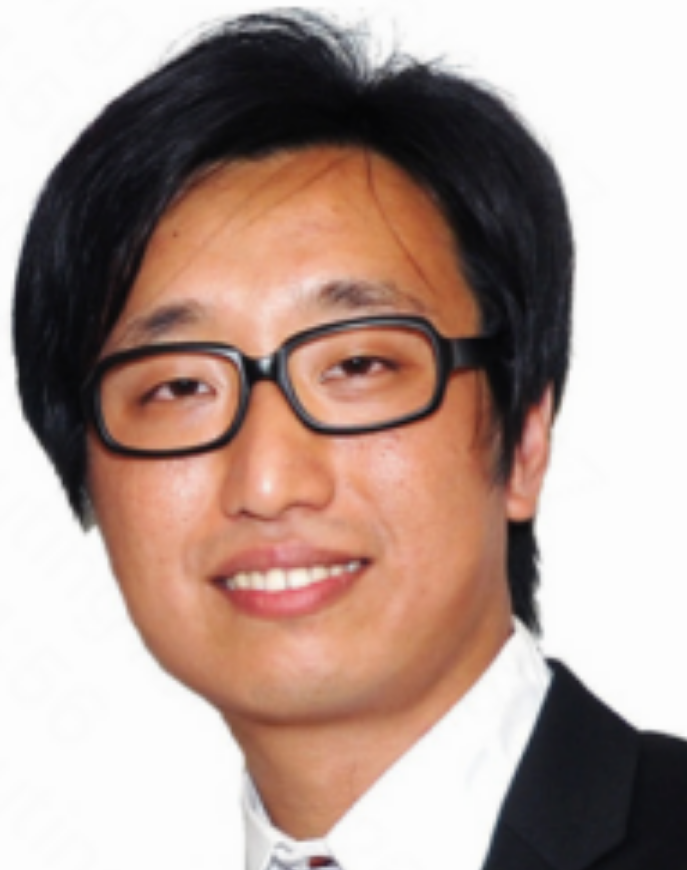}}]{Linghe Kong}
    (Fellow, IEEE) received the B.Eng. degree from Xidian University in
    2005, the master's degree from Telecom SudParis in 2007, and the Ph.D.
    degree from Shanghai Jiao Tong University in 2012. He was a
    Post-Doctoral Researcher with Columbia University, McGill University,
    and Singapore University of Technology and Design. He is currently a
    Research Professor with Shanghai Jiao Tong University. His research
    interests include satellite networking, edge and cloud computing, and
    the Internet of Things.
    \end{IEEEbiography}
  }{
    \begin{IEEEbiographynophoto}{Linghe Kong}
    (Fellow, IEEE) received the B.Eng. degree from Xidian University in
    2005, the master's degree from Telecom SudParis in 2007, and the Ph.D.
    degree from Shanghai Jiao Tong University in 2012. He was a
    Post-Doctoral Researcher with Columbia University, McGill University,
    and Singapore University of Technology and Design. He is currently a
    Research Professor with Shanghai Jiao Tong University. His research
    interests include satellite networking, edge and cloud computing, and
    the Internet of Things.
    \end{IEEEbiographynophoto}
  }
}
\newcommand{\lichaosunbiography}{
  \IfFileExists{lichao-sun.png}{
    \begin{IEEEbiography}[{\includegraphics[width=1in,height=1.25in,clip,keepaspectratio]{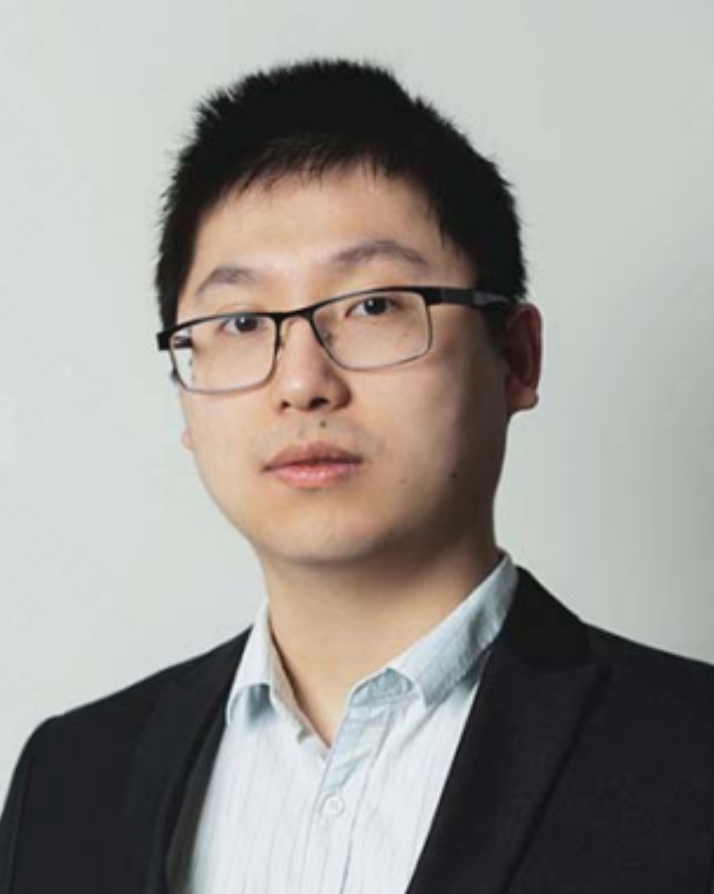}}]{Lichao Sun}
    received the Ph.D. degree from the University of Illinois at Chicago,
    Chicago, IL, USA, in 2020. He is currently an Assistant Professor with
    Lehigh University, Bethlehem, PA, USA. His research interests include
    federated learning, reinforcement learning, and deep learning. He has
    studied various computer vision, natural language processing, and graph
    applications. He has published more than 40 papers in top-tier
    conferences and journals, such as NeurIPS, KDD, AAAI, IJCAI, ICLR, CCS,
    USENIX Security, IEEE Transactions on Industrial Informatics, IEEE
    Transactions on Mobile Computing, and IEEE Transactions on Neural
    Networks and Learning Systems.
    \end{IEEEbiography}
  }{
    \begin{IEEEbiographynophoto}{Lichao Sun}
    received the Ph.D. degree from the University of Illinois at Chicago,
    Chicago, IL, USA, in 2020. He is currently an Assistant Professor with
    Lehigh University, Bethlehem, PA, USA. His research interests include
    federated learning, reinforcement learning, and deep learning. He has
    studied various computer vision, natural language processing, and graph
    applications. He has published more than 40 papers in top-tier
    conferences and journals, such as NeurIPS, KDD, AAAI, IJCAI, ICLR, CCS,
    USENIX Security, IEEE Transactions on Industrial Informatics, IEEE
    Transactions on Mobile Computing, and IEEE Transactions on Neural
    Networks and Learning Systems.
    \end{IEEEbiographynophoto}
  }
}
\newcommand{\weiranhuangbiography}{
  \IfFileExists{weiran-huang.png}{
    \begin{IEEEbiography}[{\includegraphics[width=1in,height=1.25in,clip,keepaspectratio]{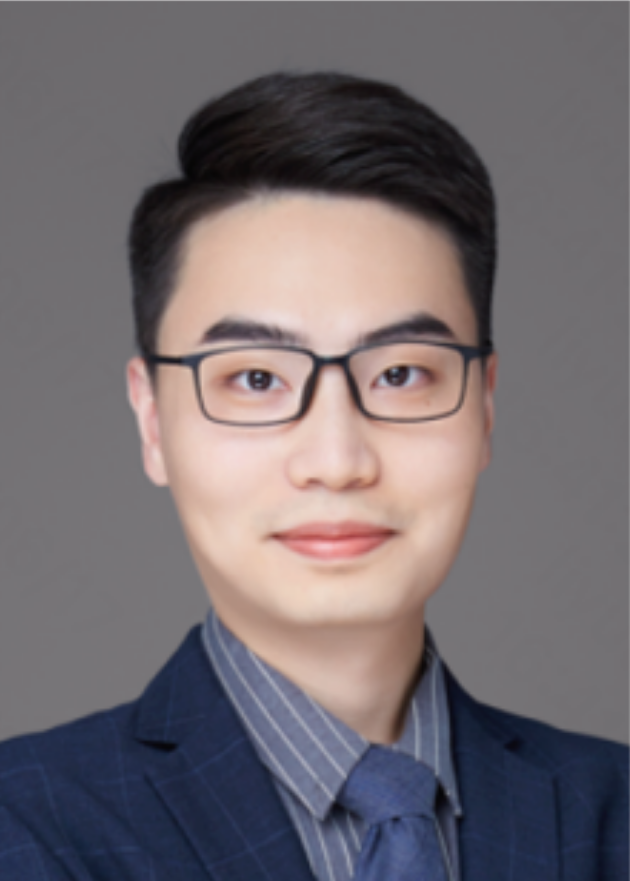}}]{Weiran Huang}
    is an Associate Professor at Shanghai Jiao Tong University and Shanghai
    Innovation Institute. His research focuses on machine learning theory,
    self-supervised learning, and few-shot learning. His work has been
    published in top-tier AI conferences, including ICML, NeurIPS, ICLR,
    CVPR, and ICCV.
    \end{IEEEbiography}
  }{
    \begin{IEEEbiographynophoto}{Weiran Huang}
    is an Associate Professor at Shanghai Jiao Tong University and Shanghai
    Innovation Institute. His research focuses on machine learning theory,
    self-supervised learning, and few-shot learning. His work has been
    published in top-tier AI conferences, including ICML, NeurIPS, ICLR,
    CVPR, and ICCV.
    \end{IEEEbiographynophoto}
  }
}
\newcommand{\yutinglibiography}{
  \IfFileExists{yuting-li.jpg}{
    \begin{IEEEbiography}[{\includegraphics[width=1in,height=1.25in,clip,keepaspectratio]{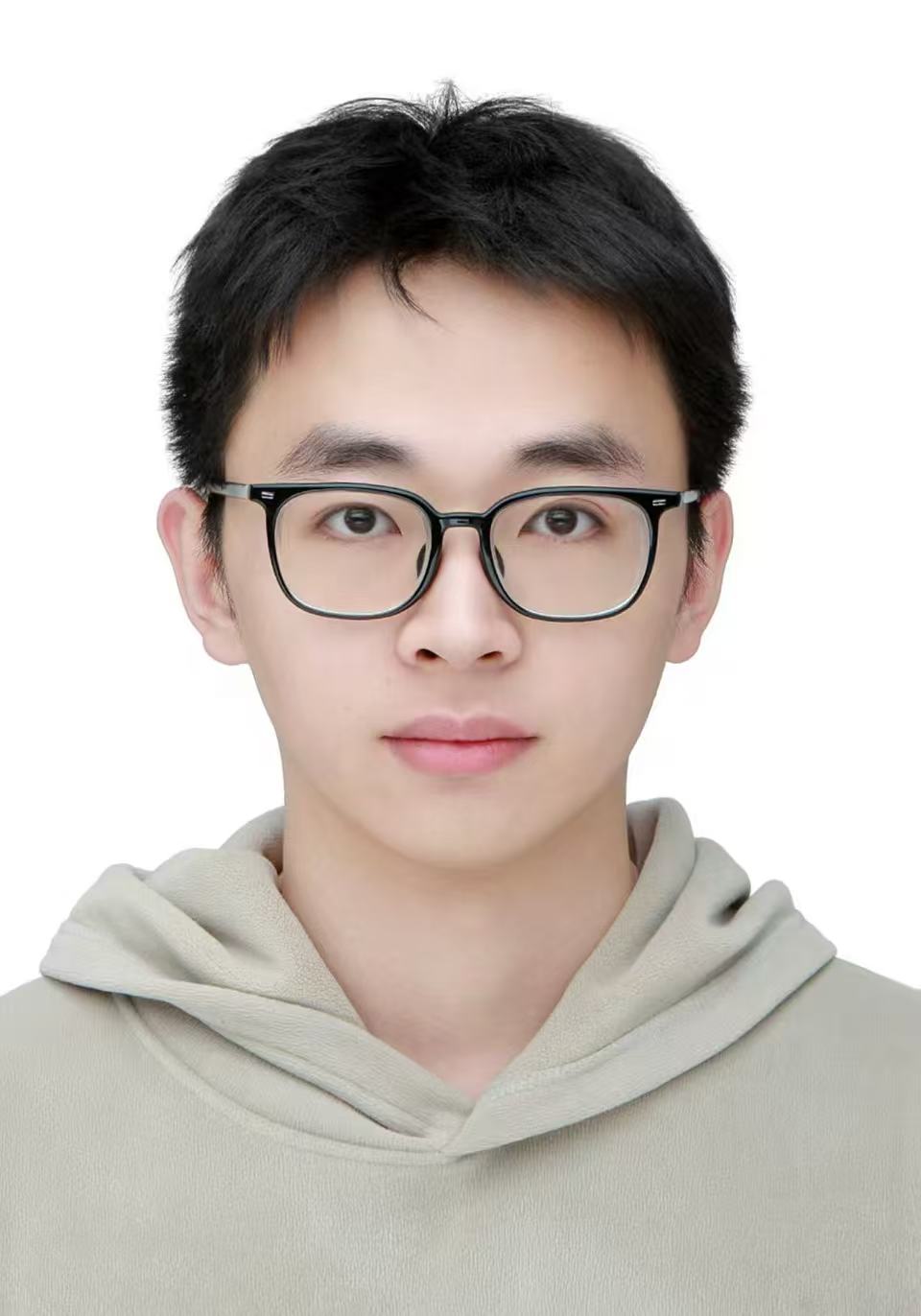}}]{Yuting Li}
    is currently a Ph.D. student in the School of Computer Science,
    Shanghai Jiao Tong University, Shanghai, China. His research focuses on
    continual learning and multi-modal learning. His work has been published
    in top-tier AI conferences, including CVPR, NeurIPS, and ICLR.
    \end{IEEEbiography}
  }{
    \begin{IEEEbiographynophoto}{Yuting Li}
    is currently a Ph.D. student in the School of Computer Science,
    Shanghai Jiao Tong University, Shanghai, China. His research focuses on
    continual learning and multi-modal learning. His work has been published
    in top-tier AI conferences, including CVPR, NeurIPS, and ICLR.
    \end{IEEEbiographynophoto}
  }
}
\newcommand{\yexinlibiography}{
  \IfFileExists{yexin-li.jpg}{
    \begin{IEEEbiography}[{\includegraphics[width=1in,height=1.25in,clip,keepaspectratio]{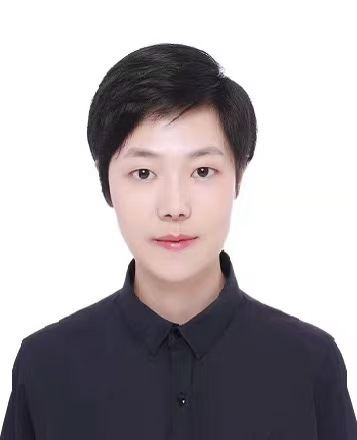}}]{Yexin Li}
    received the Ph.D. degree in Computer Science and Engineering from The
    Hong Kong University of Science and Technology in 2021. She is currently
    a Researcher with the State Key Laboratory of General Artificial
    Intelligence, Beijing Institute for General Artificial Intelligence
    (BIGAI). Before joining BIGAI, she worked as an Algorithm Engineer at
    JD.com. She also interned with the Urban Computing Group at Microsoft
    Research Asia during her doctoral studies. Her research interests include
    large language models, reinforcement learning, multi-agent systems, and
    spatio-temporal data mining.
    \end{IEEEbiography}
  }{
    \begin{IEEEbiographynophoto}{Yexin Li}
    received the Ph.D. degree in Computer Science and Engineering from The
    Hong Kong University of Science and Technology in 2021. She is currently
    a Researcher with the State Key Laboratory of General Artificial
    Intelligence, Beijing Institute for General Artificial Intelligence
    (BIGAI). Before joining BIGAI, she worked as an Algorithm Engineer at
    JD.com. She also interned with the Urban Computing Group at Microsoft
    Research Asia during her doctoral studies. Her research interests include
    large language models, reinforcement learning, multi-agent systems, and
    spatio-temporal data mining.
    \end{IEEEbiographynophoto}
  }
}
\yutinglibiography
\weihangfangbiography
\haoyuangaobiography
\linghekongbiography
\yexinlibiography
\lichaosunbiography
\weiranhuangbiography
\clearpage
\appendices
\section{Supplementary Material}
\setcounter{equation}{0}
\renewcommand{\theequation}{A\arabic{equation}}
\renewcommand{\theHequation}{appendix.A.\arabic{equation}}
\setcounter{table}{0}
\renewcommand{\thetable}{A\Roman{table}}
\renewcommand{\theHtable}{appendix.A.\arabic{table}}
\setcounter{figure}{0}
\renewcommand{\thefigure}{A\arabic{figure}}
\renewcommand{\theHfigure}{appendix.A.\arabic{figure}}
\subsection{Detailed Calculation of the Trainable Parameters about BETA and Other Baselines}
\label{sec:parms}
To ensure a fair and rigorous comparison, we analyze the trainable parameters of all methods under a unified criterion. We define \textbf{trainable parameters} as the total number of degrees of freedom that are actively updated via gradient-based or derivative-free optimization during the adaptation of a specific task. We explicitly exclude all frozen parameters of the pretrained backbone (e.g., CLIP encoders). Let $D$ denote the embedding dimension ($D=512$ for CLIP ViT-B/16) and $n_{\text{cls}}$ the number of classes.

\noindent\textbf{BETA (Ours).} The design of BETA centers on textual prototype optimization as the sole trainable substrate. A distinguishing feature of BETA is that its \textit{active} trainable parameters are \textbf{non-cumulative}; at any given training stage, only the prototypes of the current task are optimized, while past and future prototypes do not participate in the gradient calculation. 
To provide a representative value for our benchmarking, we report the parameter count based on the average class count ($n_{\text{avg}} \approx 100$) across the ten datasets in our stream (e.g., Caltech101 has 101 classes, Flowers102 has 102). Thus, the active trainable parameter count is calculated as:
\begin{equation}
    N_{\text{BETA}} = n_{\text{avg}} \times D = 100 \times 512 = 51,200 \approx \mathbf{0.05 \text{ M}}.
\end{equation}
While these optimized prototypes are stored for global task-agnostic inference, their cumulative storage remains negligible. For the full 10-dataset sequence ($\sim$1,000 classes) with CLIP ViT-B/16 ($D=512$), storing class prototypes and spherical LDR summaries requires roughly $1{,}000\times D + 1{,}000\times4\times(D+2)$ scalar values. Under mixed-precision storage, this is approximately \textbf{5 MB}. This demonstrates superior efficiency for edge-cloud deployment compared to methods that require cumulative parameter updates.

\noindent\textbf{CBBT (Tip-Adapter).} The trainable components consist of the cache adapter weights $W \in \mathbb{R}^{D \times N_{\text{cache}}}$ and context vectors. Under our configuration ($N_{\text{CTX}}=1$, CSC=False), the prompt parameters are negligible ($\approx 0.0005$ M). However, the Tip-Adapter contributes $N_{\text{tip}} = D \cdot N_{\text{cache}}$. For Caltech101 in full-shot ($N_{\text{cache}}=4,128$), $N_{\text{total}} \approx \mathbf{2.11 \text{ M}}$. In 16-shot ($N_{\text{cache}}=1,600$), $N_{\text{total}} \approx \mathbf{0.82 \text{ M}}$.

\noindent\textbf{LFA (Linear Feature Alignment).} Adaptation learns a single linear alignment matrix $T \in \mathbb{R}^{D \times D}$ to map visual features into the text-feature space. With $D=512$, the parameter count is:
\begin{equation}
    N_{\text{LFA}} = 512^2 = 262,144 \approx \mathbf{0.26 \text{ M}}.
\end{equation}

\noindent\textbf{BlackVIP.} The updated component is the Coordinator’s decoder (\textit{DecoderManual}) that generates input-dependent visual prompts. Under the default configuration, this decoder (including all layers and trigger vectors) contains approximately \textbf{0.01 M} trainable parameters.

\noindent\textbf{LCS (Black-Box Forgetting).} Adaptation is performed over a latent vector $z$ which concatenates shared and per-token components. Using the default settings ($n_{\text{ctx}}=16$, $d_{\text{ulc}}=100$, $d_{\text{slc}}=400$), the dimension $|z| = 2,000 \approx \mathbf{0.002 \text{ M}}$.

\noindent\textbf{Primal-RAIL.} Unlike BETA, Primal-RAIL analytically updates a global classifier head $W_{\text{ana}} \in \mathbb{R}^{C_{\text{total}} \times d_{\text{hid}}}$, where $C_{\text{total}}$ is the cumulative number of classes encountered so far. Under the default configuration ($d_{\text{hid}}=15,000$ and $C_{\text{total}}=1,201$), the count is:
\begin{equation}
    N_{\text{RAIL}} = 1,201 \times 15,000 \approx \mathbf{18.02 \text{ M}}.
\end{equation}
Unlike BETA’s constant overhead, Primal-RAIL's parameters scale linearly with the growth of the global label space.
\subsection{Detailed Inference Latency Analysis}
\label{sec:appen_inf}
We measure inference-time overhead in a realistic deployment protocol where the system must output \emph{post-update} predictions for the same test input. Concretely, each test batch incurs two forward passes: (i) an adaptation forward used to compute a test-time update, followed by (ii) a prediction forward executed with the updated state to produce the final logits. Experiments are conducted on Linux with Python 3.9.7 and PyTorch 2.8.0+cu128, using CUDA on an NVIDIA GeForce RTX 4090. We report median wall-clock latency with warmup; for CUDA we synchronize to ensure kernel completion before timing.
We compare \textbf{TTPA (ours)} against \textbf{TENT}~\cite{wang2020tent} and \textbf{CoTTA}~\cite{wang2022continual}. TTPA keeps the CLIP backbone frozen and adapts only a lightweight prototype/text-feature bank via at most one SGD step gated by confidence, so gradients are confined to a small $\text{num\_classes}\times d$ tensor rather than flowing through the ViT image encoder. In contrast, TENT updates normalization affine parameters by entropy minimization, which requires backpropagation through substantial parts of the network and is therefore significantly more expensive. CoTTA further increases compute by combining a teacher--student scheme with EMA teacher updates and test-time augmentation; with the default setting $\texttt{aug}=32$, the adaptation cost scales roughly with the number of augmented views in addition to teacher inference and EMA bookkeeping.
On CUDA with batch size 1 and CoTTA default $\texttt{aug}=32$, the measured single-pass latencies are 4.95\,ms for the baseline forward, 6.67\,ms for a TTPA adaptation step, 24.99\,ms for a TENT adaptation step, and 140.35\,ms for a CoTTA adaptation step. Under the two-forward protocol, the end-to-end latency is well approximated by
\[
T_{\text{2-forward}} \approx T_{\text{baseline-forward}} + T_{\text{adapt-step}},
\]
yielding 11.61\,ms (2.35$\times$) for TTPA, 29.94\,ms (6.05$\times$) for TENT, and 145.29\,ms (29.36$\times$) for CoTTA. These results highlight that TTPA achieves the best latency--adaptation trade-off by restricting updates to a small prototype/text bank, whereas TENT and especially CoTTA incur substantially higher inference-time costs due to heavy backpropagation and/or repeated multi-view forward passes.
\subsection{Additional Implementation Details}
\label{sec:appendix}
We implement all methods in PyTorch with mixed precision. We use a OneCycle learning-rate schedule and optionally enable deterministic execution with a fixed random seed. We construct class prompts using dataset-specific templates following CoOp and initialize class embeddings with the frozen text encoder. We follow standard CLIP preprocessing. All images are converted to RGB and normalized using CLIP mean and standard deviation. We apply RandomResizedCrop and RandomHorizontalFlip for training, and Resize with CenterCrop for evaluation; EuroSAT uses CenterCrop for training as well. For latent distribution replay, we encode all training images with the frozen image encoder after each task and store mixture means, spherical variances, and component weights as the replay state. During training on future tasks, we sample one pseudo-embedding per mixture component per iteration, append them to the current minibatch, and weight them by mixture weights and a replay coefficient. We optionally enable prototype alignment loss and feature mixup. For test-time prototype alignment, we optimize the class embedding parameters with SGD and momentum $0.9$ under a confidence threshold $\delta=0.95$.
We report the learning rate and the number of training epochs used for each dataset in the Black-CL stream in Table~\ref{tab:per_dataset_hparams}.
\begin{table}[ht]
\centering
\caption{Per-dataset learning rates and training epochs used for 16-shot and full-shot settings.}
\label{tab:per_dataset_hparams}
\small
\setlength{\tabcolsep}{5pt}
\renewcommand{\arraystretch}{1.15}
\begin{tabular}{@{}l rr|rr@{}}
\hline
 & \multicolumn{2}{c|}{16-shot} & \multicolumn{2}{c@{}}{Full-shot} \\
\cline{2-3}\cline{4-5}
Dataset & LR & Epochs & LR & Epochs \\
\hline
Caltech101 & 0.005 & 100 & 0.0005 & 200 \\
OxfordPets & 0.0005 & 200 & 0.0005 & 100 \\
StanfordCars & 0.001 & 100 & 0.001 & 100 \\
Flowers102 & 0.005 & 200 & 0.001 & 200 \\
Food101 & 0.0001 & 200 & 0.0002 & 200 \\
FGVCAircraft & 0.005 & 200 & 0.02 & 100 \\
SUN397 & 0.0005 & 100 & 0.0005 & 100 \\
DTD & 0.005 & 100 & 0.001 & 200 \\
EuroSAT & 0.005 & 200 & 0.005 & 200 \\
UCF101 & 0.0005 & 200 & 0.001 & 100 \\
\hline
\end{tabular}
\end{table}
\subsection{Robustness to Dataset Order}
\label{sec:dataset_order}
To evaluate sensitivity to the task sequence, we repeat the full-shot
experiment using the alternative dataset order adopted by
LADA~\cite{luo2025ladascalablelabelspecificclip} and
Primal-RAIL~\cite{xu2024advancing}: StanfordCars, FGVCAircraft, OxfordPets,
Food101, SUN397, UCF101, Flowers102, DTD, Caltech101, and EuroSAT. We keep all hyperparameters fixed to those
used for the original order. As shown in Table~\ref{tab:dataset_order}, BETA
continues to deliver compelling performance across Transfer, Average, and Last
Accuracy, demonstrating its robustness to changes in dataset order.
\begin{table*}[t]
\centering
\caption{\textbf{Full-shot performance under an alternative dataset order.}
We report the complete evaluation matrix for the order
StanfordCars, FGVCAircraft, OxfordPets, Food101, SUN397, UCF101, Flowers102,
DTD, Caltech101, and EuroSAT. Changes in the mean metrics relative to the
original order are shown in parentheses.}
\label{tab:dataset_order}
\scriptsize
\setlength{\tabcolsep}{3.2pt}
\renewcommand{\arraystretch}{1.08}
\begin{adjustbox}{width=\textwidth}
\begin{tabular}{l cccccccccc c}
\toprule
\textbf{Training Stage / Metric} &
\textbf{Sta.} & \textbf{FGV.} & \textbf{Oxf.} & \textbf{Foo.} &
\textbf{SUN} & \textbf{UCF} & \textbf{Flo.} & \textbf{DTD} &
\textbf{Cal.} & \textbf{Eur.} & \textbf{Mean} \\
\midrule
StanfordCars & 87.0 & 24.7 & 89.1 & 77.0 & 61.9 & 66.3 & 71.3 & 37.1 & 75.3 & 48.5 & 63.8 \\
FGVCAircraft & 87.0 & 53.8 & 89.1 & 77.1 & 62.2 & 66.3 & 71.3 & 37.1 & 76.1 & 48.5 & 66.8 \\
OxfordPets & 87.0 & 53.8 & 93.9 & 77.0 & 62.1 & 66.3 & 71.3 & 36.9 & 75.5 & 48.6 & 67.2 \\
Food101 & 87.0 & 53.8 & 93.9 & 89.3 & 62.0 & 66.6 & 71.1 & 36.6 & 75.6 & 48.4 & 68.4 \\
SUN397 & 87.0 & 53.9 & 93.9 & 89.5 & 75.6 & 65.5 & 71.0 & 37.1 & 75.1 & 48.2 & 69.7 \\
UCF101 & 87.0 & 53.9 & 94.1 & 89.5 & 77.0 & 86.4 & 71.0 & 37.0 & 76.9 & 48.2 & 72.1 \\
Flowers102 & 87.0 & 53.9 & 94.1 & 89.5 & 77.1 & 86.4 & 98.7 & 37.8 & 77.3 & 48.1 & 75.0 \\
DTD & 87.0 & 53.9 & 94.1 & 89.5 & 77.1 & 86.4 & 98.7 & 75.3 & 77.0 & 47.6 & 78.7 \\
Caltech101 & 87.0 & 54.1 & 94.3 & 89.8 & 77.6 & 86.4 & 98.9 & 75.7 & 94.9 & 45.9 & 80.5 \\
EuroSAT & 87.0 & 54.1 & 94.3 & 89.8 & 77.6 & 86.4 & 98.9 & 75.7 & 94.9 & 95.9 & 85.5 \\
\midrule
Transfer Accuracy & -- & 24.7 & 89.1 & 77.0 & 62.0 & 66.2 & 71.2 & 37.1 & 76.1 & 48.0 & \textbf{61.3} {\small (+1.5)} \\
Average Accuracy & 87.0 & 51.0 & 93.1 & 85.8 & 71.0 & 76.3 & 82.2 & 48.6 & 79.9 & 52.8 & \textbf{72.8} {\small (+0.3)} \\
Last Accuracy & 87.0 & 54.1 & 94.3 & 89.8 & 77.6 & 86.4 & 98.9 & 75.7 & 94.9 & 95.9 & \textbf{85.5} {\small (+0.1)} \\
\bottomrule
\end{tabular}
\end{adjustbox}
\end{table*}
\subsection{16-Shot Experimental Results}
\label{sec:16-shot results-bcl}
The main paper reports the full-shot experimental results matching the 16-shot experiments in Table~\ref{tab:16_shot_results}, Table~\ref{tab:cross_dataset_transfer}, and Table~\ref{tab:backbone_comparison_6rows}. All core experimental configurations are kept consistent, with only the per-class sample size, learning rate and training epochs adjusted for the full-shot setting. These results further verify our proposed method's effectiveness and robustness in data-sufficient conditions.
\begin{table*}[t]
\centering
\caption{\textbf{Main results on the 16-shot black-box continual learning task stream}. We evaluate performance across 10 diverse datasets. We report \textbf{Transfer Accuracy}, \textbf{Average Accuracy}, and \textbf{Last Accuracy}. \textbf{Bold} indicates the best performance among black-box methods. The main paper provides the corresponding full-shot results.}
\label{tab:16_shot_results}
\begin{adjustbox}{width=\textwidth}
\begin{tabular}{l c cccccccccc >{\columncolor{gray!10}}c}
\toprule
& &\multicolumn{11}{c}{Continual Learning Task Stream $\mathcal{S}$} \\
 \cmidrule(lr){3-13}
& Params & \rotatebox{90}{Caltech} & \rotatebox{90}{Pets} & \rotatebox{90}{Cars} & \rotatebox{90}{Flowers} & \rotatebox{90}{Food} & \rotatebox{90}{Aircraft} & \rotatebox{90}{SUN397} & \rotatebox{90}{DTD} & \rotatebox{90}{EuroSAT} & \rotatebox{90}{UCF101} & \cellcolor{white}\rotatebox{90}{\textit{Average}} \\
\midrule
Zero-shot & -- & 72.4 & 88.6 & 65.5 & 70.7 & 85.3 & 24.8 & 59.1 & 33.7 & 45.2 & 64.9 & 61.0 \\ 
\midrule
& &\multicolumn{11}{c}{Transfer Accuracy} \\
 \cmidrule(lr){3-13}
CBBT~\cite{guo2023black}  & 0.8 M & -- & 86.8 & 63.5 & 67.8 & 81.2 & 20.9 & 56.4 & 30.8 & 41.5 & 61.5 & 56.7 \\
LFA~\cite{ouali2023black}   & 0.3 M & -- & 86.4 & 62.8 & 66.5 & 80.4 & 19.8 & 54.9 & 29.5 & 40.2 & 60.8 & 55.7 \\
BlackVIP~\cite{oh2023blackvip}   & 0.01 M & -- & 84.5 & 61.5 & 65.2 & 78.5 & 18.7 & 53.4 & 28.6 & 39.4 & 59.2 & 54.3 \\
LCS~\cite{kuwana2024black}   & 0.002 M & -- & 84.1 & 60.5 & 64.2 & 77.8 & 17.5 & 52.8 & 27.2 & 38.6 & 58.4 & 53.5 \\
RAIL~\cite{xu2024advancing}     & 18.0 M & -- & 88.6 & \textbf{65.5} & \textbf{70.7} & \textbf{85.3} & \textbf{24.8} & 59.1 & 33.7 & 45.2 & 64.9 & 59.7 \\ 
\textbf{BETA (Ours)}   & \textbf{0.05 M} & -- & \textbf{88.6} & 62.8 & 69.7 &  \textbf{76.2} & 23.9  & \textbf{61.9} & \textbf{38.0} & \textbf{49.5} & \textbf{67.2} & \textbf{59.8}  \\
\midrule
& &\multicolumn{11}{c}{Average Accuracy} \\
 \cmidrule(lr){3-13}
CBBT~\cite{guo2023black}  & 0.8 M & 84.1 & 88.4 & 71.8 & 80.5 & 81.5 & 29.8 & 61.5 & 38.5 & 44.2 & 60.5 & 64.1 \\
LFA~\cite{ouali2023black}   & 0.3 M & 83.2 & 87.5 & 70.4 & 79.2 & 80.2 & 28.5 & 59.8 & 37.2 & 43.1 & 59.4 & 62.9 \\
BlackVIP~\cite{oh2023blackvip}   & 0.01 M & 81.1 & 85.8 & 68.5 & 77.4 & 78.8 & 26.7 & 57.9 & 35.5 & 41.2 & 57.2 & 61.0 \\
LCS~\cite{kuwana2024black}   & 0.002 M & 80.2 & 84.9 & 67.5 & 76.5 & 77.5 & 25.4 & 56.5 & 34.1 & 40.2 & 56.1 & 59.9 \\
RAIL~\cite{xu2024advancing}     & 18.0 M & 82.7 & 91.9 & 78.1 & 86.7 & 85.2 & 33.9 & 64.7 & 43.7 & 47.5 & 62.6 & 67.7  \\
\textbf{BETA (Ours)}   & \textbf{0.05 M} & \textbf{89.9} & \textbf{92.1} & \textbf{79.7} & \textbf{88.9} & \textbf{82.5} & \textbf{35.4} & \textbf{66.3} & \textbf{46.8} & \textbf{57.3} & \textbf{68.8} & \textbf{70.8}  \\
\midrule
& &\multicolumn{11}{c}{Last Accuracy} \\
 \cmidrule(lr){3-13}
CBBT~\cite{guo2023black}  & 0.8 M & 81.4 & 80.1 & 67.5 & 82.4 & 78.8 & 26.4 & 57.5 & 34.1 & 49.2 & 68.4 & 62.6 \\
LFA~\cite{ouali2023black}   & 0.3 M & 80.8 & 79.4 & 66.5 & 81.8 & 77.5 & 25.7 & 56.4 & 33.1 & 48.4 & 67.5 & 61.7 \\
BlackVIP~\cite{oh2023blackvip}   & 0.01 M & 78.5 & 77.8 & 64.5 & 79.8 & 75.4 & 23.8 & 54.4 & 31.2 & 46.5 & 65.4 & 59.7 \\
LCS~\cite{kuwana2024black}   & 0.002 M & 77.5 & 76.5 & 63.2 & 78.5 & 74.2 & 22.5 & 53.2 & 30.1 & 45.2 & 64.1 & 58.7 \\
RAIL~\cite{xu2024advancing}     & 18.0 M & \textbf{94.1} & 92.6 & 81.0& 96.1 & \textbf{86.8} & 43.9 & \textbf{75.0} & \textbf{68.3} & 85.9 & 81.9 & 80.6 \\
\textbf{BETA (Ours)}   & \textbf{0.05 M} & 93.4  & \textbf{92.8} & \textbf{83.9} & \textbf{97.1} & \textbf{86.8} & \textbf{46.9} & 74.7 & 67.8 & \textbf{88.4} & \textbf{83.1} & \textbf{81.5}  \\
\bottomrule
\end{tabular}
\end{adjustbox}
\end{table*}
\begin{table*}[t]
\centering
\caption{\textbf{Comparison with state-of-the-art white-box methods on the 16-shot X-TAIL task stream.} We report \textbf{Transfer Accuracy}, \textbf{Average Accuracy}, and \textbf{Last Accuracy} across ten datasets. Unlike baselines that require full model access ($\times$), \textbf{BETA} operates under strict black-box constraints ($\surd$) and utilizes significantly fewer trainable parameters (only 0.05 M). \textbf{Bold} indicates the best performance among all methods. The main paper provides the corresponding full-shot results.}
\label{tab:cross_dataset_transfer}
\begin{adjustbox}{width=\textwidth}
\begin{tabular}{l cc cccccccccc >{\columncolor{gray!10}}c}
\toprule
& & &\multicolumn{11}{c}{Continual Learning Task Stream $\mathcal{S}$} \\
 \cmidrule(lr){4-14}
& Black-Box ? & Params & \rotatebox{90}{Aircraft} & \rotatebox{90}{Caltech} & \rotatebox{90}{DTD} & \rotatebox{90}{EuroSAT} & \rotatebox{90}{Flowers} & \rotatebox{90}{Food} & \rotatebox{90}{MNIST} & \rotatebox{90}{Pets} & \rotatebox{90}{Cars} & \rotatebox{90}{Sun397} & \cellcolor{white}\rotatebox{90}{\textit{Average}} \\
\midrule
Zero-shot & -- & -- & 23.8 & 74.3 &36.4 &37.4 &64.1& 83.4 &43.9 &87.8 &65.5 &60.8 &57.7 \\ 
\midrule
& & &\multicolumn{11}{c}{Transfer Accuracy} \\
 \cmidrule(lr){4-14}
LwF~\cite{li2017learningforgetting}    & $\times$ & 149.6 M & -- & 66.6 & 26.9 &19.5 &51.0 &78.4 &26.6 &68.9 &35.5 &56.1 &47.7\\
WiSE-FT~\cite{wortsman2022robust}    & $\times$ & 149.6 M & -- &70.1 & 31.9& 25.3 &56.3& 79.8& 29.9& 74.9 &45.6 &56.8 &52.3  \\
ZSCL~\cite{zheng2023preventing}  & $\times$ & 149.6 M & -- & 73.3 & 32.6 & 36.8 & 62.1 &83.8 &42.1 &83.6 &56.5 &60.2 &59.0  \\
MoE-Adapters~\cite{yu2024boosting}    & $\times$ & 59.8 M& -- &71.0 & 34.9 & 19.2 & 63.0 & \textbf{86.6} & 20.0 & 87.2 & 63.7 & 58.6 & 56.0  \\
LADA~\cite{luo2025ladascalablelabelspecificclip} & $\times$ & 9.0 M& -- &75.0 &\textbf{36.1} &35.9& 66.3 &83.7 &42.1 &88.0 &\textbf{65.3} &61.4 &61.5 \\
\textbf{BETA (Ours)}    & $\surd$ & \textbf{0.05 M} & -- & \textbf{75.9} & 34.7 & \textbf{46.6} & \textbf{70.1} & 75.2 & \textbf{43.2} & \textbf{89.2} & 62.7 & \textbf{64.5} & \textbf{62.5}\\
\midrule
& & &\multicolumn{11}{c}{Average Accuracy} \\
 \cmidrule(lr){4-14}
LwF~\cite{li2017learningforgetting}    & $\times$ & 149.6 M & 24.7 & 79.7 & 38.3 &36.9 &63.9& 81.0 &36.5 &71.9 &42.7 &56.7& 53.2 \\
WiSE-FT~\cite{wortsman2022robust}    & $\times$ & 149.6 M & 27.1 & 76.5 & 40.9 & 31.3 & 68.7 & 81.6 &31.4 &74.7 &51.7& 58.4 & 54.2  \\
ZSCL~\cite{zheng2023preventing}  & $\times$ & 149.6 M & 36.0 & 75.0 &40.7 &40.5 &71.0 &85.3 &46.3 &83.3 &60.7 & 61.5 & 60.0  \\
MoE-Adapters~\cite{yu2024boosting}    & $\times$ & 59.8 M& 43.6 & 77.9 & 52.1 & 34.7 & 75.9 &\textbf{86.3} &45.2 &87.4 &66.6 &60.2 &63.0 \\
LADA~\cite{luo2025ladascalablelabelspecificclip} & $\times$ & 9.0 M& \textbf{49.1} & \textbf{91.0}& 61.3 &71.6 &84.4 &85.0& 62.8 &89.7 &\textbf{69.2} &62.9 & \textbf{72.7} \\
\textbf{BETA (Ours)}    & $\surd$ & \textbf{0.05 M} & 46.5 & 85.2 & \textbf{62.5} & \textbf{75.8} & \textbf{86.2} & 80.7 & \textbf{62.8} & \textbf{90.2} & 66.9 & \textbf{65.6}&72.2 \\
\midrule
& & &\multicolumn{11}{c}{Last Accuracy} \\
 \cmidrule(lr){4-14}
LwF~\cite{li2017learningforgetting}    & $\times$ & 149.6 M & 20.9 & 83.1& 47.5& 38.2& 75.5& 84.7 &50.1 &78.0& 75.8& 74.6& 62.8  \\
WiSE-FT~\cite{wortsman2022robust}    & $\times$ & 149.6 M & 21.8 &76.8 &42.9 &20.8 &77.5 &84.9& 30.7 &76.6 &75.8 &72.5 &58.0  \\
ZSCL~\cite{zheng2023preventing}  & $\times$ & 149.6 M & 33.1 & 75.3 & 43.5 & 35.2 & 74.6 & \textbf{87.4} & 50.4 &84.2 &77.3 &73.4 & 63.4  \\
MoE-Adapters~\cite{yu2024boosting}    & $\times$ & 59.8 M & 43.2 & 78.7 & 57.6 & 32.8 & 79.4 &86.0 &86.7 &87.8 &78.2 &74.2 &70.5 \\
LADA~\cite{luo2025ladascalablelabelspecificclip} & $\times$ & 9.0 M &\textbf{49.6} & \textbf{93.7} & 69.3 & 86.9 &96.7 &86.9 &\textbf{93.8} &\textbf{93.7} &\textbf{84.6} &\textbf{76.0} & \textbf{83.1} \\
\textbf{BETA (Ours)}    & $\surd$ & \textbf{0.05 M} & 46.7 & 88.0 & \textbf{71.0} & \textbf{88.4} & \textbf{96.9} & 86.2 & 92.0 & 92.8 & 83.7 & 75.0&82.1 \\
\bottomrule
\end{tabular}
\end{adjustbox}
\end{table*}
\begin{table*}[t]
\centering
\caption{\textbf{Performance gains of BETA across various backbones.} Each backbone spans multiple rows to accommodate different variants and their corresponding adaptation results. $\Delta$ denotes the gain over the Zero-shot baseline. The main paper provides the corresponding full-shot results.}
\label{tab:backbone_comparison_6rows}
\begin{adjustbox}{width=\textwidth}
\begin{tabular}{lll cccccccccc c}
\toprule
& & & \multicolumn{11}{c}{\textbf{Continual Learning Task Stream $\mathcal{S}$}} \\
\cmidrule(lr){4-14}
\textbf{Backbone} & \textbf{Variant} & \textbf{Method} & \rotatebox{90}{Caltech} & \rotatebox{90}{Pets} & \rotatebox{90}{Cars} & \rotatebox{90}{Flowers} & \rotatebox{90}{Food} & \rotatebox{90}{Aircraft} & \rotatebox{90}{SUN397} & \rotatebox{90}{DTD} & \rotatebox{90}{EuroSAT} & \rotatebox{90}{UCF101} & \rotatebox{90}{\textit{Average}} \\
\midrule
\multirow{6}{*}{CLIP} & \multirow{3}{*}{B/16-224} 
& Zero-shot & 72.4 & 88.6 & 65.5 & 70.7 & 85.3 & 24.8 & 59.1 & 33.7 & 45.2 & 64.9 & 61.0 \\
& & BETA (Ours) &93.4  &92.8 & 83.9 & 97.1 & 86.8 & 46.9 & 74.7 & 67.8 & 88.4 & 83.1 & 81.5  \\
& &  $\Delta$  &  \cellcolor{gray!10}\textcolor{gain}{+21.0} & \cellcolor{gray!10}\textcolor{gain}{+4.2} & \cellcolor{gray!10}\textcolor{gain}{+18.4} & \cellcolor{gray!10}\textcolor{gain}{+26.4} & \cellcolor{gray!10}\textcolor{gain}{+1.5} & \cellcolor{gray!10}\textcolor{gain}{+22.1} & \cellcolor{gray!10}\textcolor{gain}{+15.6} & \cellcolor{gray!10}\textcolor{gain}{+34.1} & \cellcolor{gray!10}\textcolor{gain}{+43.2} & \cellcolor{gray!10}\textcolor{gain}{+18.2} & \cellcolor{gray!10}\textcolor{gain}{+20.5} \\
\cmidrule(lr){2-14}
& \multirow{3}{*}{L/14-224} 
& Zero-shot & 75.7 &  93.3 &  76.8 &  79.5 &  90.3 & 32.5 & 64.4 & 41.8 & 61.0 & 73.3 & 68.9 \\
& & BETA (Ours) & 94.0 & 94.9 & 89.2 & 98.7 & 90.8 & 59.1 & 78.7 & 71.5 & 91.6 & 85.1 & 85.4 \\
& &  $\Delta$ & \cellcolor{gray!10}\textcolor{gain}{+18.3} & \cellcolor{gray!10}\textcolor{gain}{+1.6} & \cellcolor{gray!10}\textcolor{gain}{+12.4} & \cellcolor{gray!10}\textcolor{gain}{+19.2} & \cellcolor{gray!10}\textcolor{gain}{+0.5} & \cellcolor{gray!10}\textcolor{gain}{+26.6} & \cellcolor{gray!10}\textcolor{gain}{+14.3} & \cellcolor{gray!10}\textcolor{gain}{+29.7} & \cellcolor{gray!10}\textcolor{gain}{+30.6} & \cellcolor{gray!10}\textcolor{gain}{+11.8} & \cellcolor{gray!10}\textcolor{gain}{+16.5} \\
\midrule
\multirow{6}{*}{SigLIP} & \multirow{3}{*}{B/16-224} 
& Zero-shot & 85.3 & 92.8 & 89.0 & 81.7 & 88.0 & 29.4 & 65.7 & 55.1 & 31.6 & 72.2 &  69.1 \\
& & BETA (Ours) & 95.4 & 95.0 & 93.6 & 99.2 & 89.4 & 53.5 & 76.4 & 76.7 & 87.9 & 80.8 & 84.8 \\
& &  $\Delta$ & \cellcolor{gray!10}\textcolor{gain}{+10.1} & \cellcolor{gray!10}\textcolor{gain}{+2.2} & \cellcolor{gray!10}\textcolor{gain}{+4.6} & \cellcolor{gray!10}\textcolor{gain}{+17.5} & \cellcolor{gray!10}\textcolor{gain}{+1.4} & \cellcolor{gray!10}\textcolor{gain}{+24.1} & \cellcolor{gray!10}\textcolor{gain}{+10.7} & \cellcolor{gray!10}\textcolor{gain}{+21.6} & \cellcolor{gray!10}\textcolor{gain}{+56.3} & \cellcolor{gray!10}\textcolor{gain}{+8.6} & \cellcolor{gray!10}\textcolor{gain}{+15.7} \\
\cmidrule(lr){2-14}
& \multirow{3}{*}{So/14-224} 
& Zero-shot & 85.4 & 95.0 & 87.5 & 89.5 & 92.6 & 50.1 & 72.0 & 56.4 & 49.4 & 77.6 & 75.5\\
& & BETA (Ours) & 94.6 & 96.2 & 95.3 & 99.6 & 93.0 & 69.0 & 81.0 & 79.0 & 90.1 & 82.7 & 88.1 \\
& &  $\Delta$ & \cellcolor{gray!10}\textcolor{gain}{+9.2} & \cellcolor{gray!10}\textcolor{gain}{+1.2} & \cellcolor{gray!10}\textcolor{gain}{+7.8} & \cellcolor{gray!10}\textcolor{gain}{+10.1} & \cellcolor{gray!10}\textcolor{gain}{+0.4} & \cellcolor{gray!10}\textcolor{gain}{+18.9} & \cellcolor{gray!10}\textcolor{gain}{+9.0} & \cellcolor{gray!10}\textcolor{gain}{+22.6} & \cellcolor{gray!10}\textcolor{gain}{+40.7} & \cellcolor{gray!10}\textcolor{gain}{+5.1} & \cellcolor{gray!10}\textcolor{gain}{+12.6} \\
\midrule
\multirow{6}{*}{SigLIP 2} & \multirow{3}{*}{B/16-224} 
& Zero-shot & 85.8 & 92.7 & 92.5 & 81.6 & 88.8 & 28.0 & 69.4 & 55.1 & 35.8 & 66.2 & 69.6 \\
& & BETA (Ours) & 95.2 & 95.0 & 94.0 & 99.4 & 90.2 & 60.3 & 78.0 & 76.8 & 87.9 & 83.0 & 86.0 \\
& &  $\Delta$ & \cellcolor{gray!10}\textcolor{gain}{+9.4} & \cellcolor{gray!10}\textcolor{gain}{+2.3} & \cellcolor{gray!10}\textcolor{gain}{+1.5} & \cellcolor{gray!10}\textcolor{gain}{+17.8} & \cellcolor{gray!10}\textcolor{gain}{+1.4} & \cellcolor{gray!10}\textcolor{gain}{+32.3} & \cellcolor{gray!10}\textcolor{gain}{+8.6} & \cellcolor{gray!10}\textcolor{gain}{+21.7} & \cellcolor{gray!10}\textcolor{gain}{+52.1} & \cellcolor{gray!10}\textcolor{gain}{+16.8} & \cellcolor{gray!10}\textcolor{gain}{+16.4} \\
\cmidrule(lr){2-14}
& \multirow{3}{*}{So/14-224} 
& Zero-shot & 85.3 & 94.2 & 95.0 & 89.2 & 92.5 & 55.0 & 70.8 & 60.1 & 43.3 & 76.0 & 76.1 \\
& & BETA (Ours) & 95.7 & 96.2 & 95.9 & 99.8 & 93.4 & 77.3 & 81.1 & 78.5 & 87.0 & 84.4 & 88.9 \\
& &  $\Delta$ & \cellcolor{gray!10}\textcolor{gain}{+10.4} & \cellcolor{gray!10}\textcolor{gain}{+2.0} & \cellcolor{gray!10}\textcolor{gain}{+0.9} & \cellcolor{gray!10}\textcolor{gain}{+10.6} & \cellcolor{gray!10}\textcolor{gain}{+0.9} & \cellcolor{gray!10}\textcolor{gain}{+22.3} & \cellcolor{gray!10}\textcolor{gain}{+10.3} & \cellcolor{gray!10}\textcolor{gain}{+18.4} & \cellcolor{gray!10}\textcolor{gain}{+44.7} & \cellcolor{gray!10}\textcolor{gain}{+8.4} & \cellcolor{gray!10}\textcolor{gain}{+12.8} \\
\bottomrule
\end{tabular}
\end{adjustbox}
\end{table*}
\subsection{Comprehensive results of BETA Across Various Backbones}
To comprehensively validate the generalization and effectiveness of our proposed Beta method, we conduct systematic experiments across six representative vision-language backbones, spanning both 16-shot (data-scarce) and full-shot (data-sufficient) evaluation regimes. The selected backbones include: CLIP ViT-B/16 (224×224), CLIP ViT-L/14 (224×224), SigLIP Variant 1 ViT-B/16 (224×224), SigLIP Variant 1 So/14 (224×224), SigLIP Variant 2 ViT-B/16 (224×224), and SigLIP Variant 2 So/14 (224×224).
All core experimental protocols are kept consistent across these backbones to ensure fairness: we only adjust per-class sample sizes, learning rates, and training epochs to adapt to the 16-shot and full-shot settings, respectively. The following tables present the comprehensive results.
\begin{table*}[!t]
\centering
\caption{\textbf{Comprehensive results of BETA in the black-box continual learning setting on the 16-shot task stream $\mathcal{S}$ with \textbf{CLIP-B/16-224} backbone.} Each row represents the performance on every dataset of the model trained after the corresponding task. We report accuracy (\%), where the diagonal (\colorbox{diag_green}{green}) represents the performance on the current task, and the last row (\colorbox{last_red}{red}) denotes the Last Accuracy.}
\label{tab:continual_matrix1}
\begin{adjustbox}{width=0.7\textwidth}
\begin{tabular}{l cccccccccc >{\columncolor{avg_yellow}}c}
\toprule
& \rotatebox{90}{Caltech101} & \rotatebox{90}{OxfordPets} & \rotatebox{90}{StanfordCars} & \rotatebox{90}{Flowers102} & \rotatebox{90}{Food101} & \rotatebox{90}{FGVCAircraft} & \rotatebox{90}{SUN397} & \rotatebox{90}{DTD} & \rotatebox{90}{EuroSAT} & \rotatebox{90}{UCF101} & \textbf{Average} \\
\midrule
Zero-shot & 72.4 & 88.6 &65.5 &70.7 &85.3 &24.8 &59.1 &33.7 &45.2& 64.9 &61.0 \\
\midrule
Caltech101   & \cellcolor{diag_green}86.1 & 88.6&  62.8&  69.7&  76.2 & 23.9 & 61.8 & 37.8 & 50.0  & 66.8 & 62.4 \\
OxfordPets   & 86.4 & \cellcolor{diag_green}92.4 & 62.8 & 69.7&  76.2&  23.9 & 61.8 & 37.6&  50.0 & 66.7 & 62.8 \\
StanfordCars & 86.7  & 92.4 & \cellcolor{diag_green}83.9 & 69.8 & 76.2 & 23.9&  61.9 & 37.7 & 50.0 & 66.7 & 64.9 \\
Flowers102   & 87.6 & 92.3 & 83.9 & \cellcolor{diag_green}97.1 & 76.3 & 23.9 & 61.8 & 38.5 & 50.0 & 66.9 & 67.8 \\
Food101      & 88.0 & 92.3 & 83.9 & 97.1 & \cellcolor{diag_green}86.7 & 24.0 & 62.0  &37.4 & 49.7 & 66.8 & 68.8 \\
FGVCAircraft & 92.0 & 92.3 & 83.9 & 97.1 & 86.7 & \cellcolor{diag_green}46.9 & 61.9 & 37.5 & 49.9 & 66.8 & 71.5 \\
SUN397       & 92.8 & 92.5 & 83.9 & 97.1 & 86.8 & 46.9 & \cellcolor{diag_green}72.4 & 39.2 & 48.8 & 67.8 & 72.8 \\
DTD          & 92.9 & 92.5 & 83.9 & 97.1 & 86.8 & 46.9 & 72.4 & \cellcolor{diag_green}67.2 & 47.9 & 68.1 & 75.6 \\
EuroSAT      & 92.9 &  92.5 & 83.9 & 97.1 & 86.8 & 46.9 & 72.4 & 67.3 & \cellcolor{diag_green}88.2 & 68.0 & 75.6 \\ 
UCF101       & \cellcolor{last_red}93.4 & \cellcolor{last_red}92.8 & \cellcolor{last_red}83.9 & \cellcolor{last_red}97.1 & \cellcolor{last_red}86.8 & \cellcolor{last_red}46.9 & \cellcolor{last_red}74.7 & \cellcolor{last_red}67.8 & \cellcolor{last_red}88.4 & \cellcolor{last_red}83.1 & \textbf{81.5} \\
\midrule
\textbf{Average} & 89.9 & 92.1 & 79.7 & 88.9 & 82.5 & 35.4 & 66.3 & 46.8 & 57.3 & 68.8& \cellcolor{avg_yellow}\textbf{70.8} \\
\bottomrule
\end{tabular}
\end{adjustbox}
\end{table*}
\begin{table*}[!t]
\centering
\caption{\textbf{Comprehensive results of BETA in the black-box continual learning setting on the full-shot task stream $\mathcal{S}$ with \textbf{CLIP-B/16-224} backbone.} Each row represents the performance on every dataset of the model trained after the corresponding task. We report accuracy (\%), where the diagonal (\colorbox{diag_green}{green}) represents the performance on the current task, and the last row (\colorbox{last_red}{red}) denotes the Last Accuracy.}
\begin{adjustbox}{width=0.7\textwidth}
\begin{tabular}{l cccccccccc >{\columncolor{avg_yellow}}c}
\toprule
& \rotatebox{90}{Caltech101} & \rotatebox{90}{OxfordPets} & \rotatebox{90}{StanfordCars} & \rotatebox{90}{Flowers102} & \rotatebox{90}{Food101} & \rotatebox{90}{FGVCAircraft} & \rotatebox{90}{SUN397} & \rotatebox{90}{DTD} & \rotatebox{90}{EuroSAT} & \rotatebox{90}{UCF101} & \textbf{Average} \\
\midrule
Zero-shot & 72.4 & 88.6 &65.5 &70.7 &85.3 &24.8 &59.1 &33.7 &45.2& 64.9 &61.0 \\
\midrule
Caltech101   & \cellcolor{diag_green}88.1 &  88.5 & 62.8 & 69.1 & 76.5 & 23.9 & 61.7 & 37.8 & 50.0 & 66.5 & 62.5 \\
OxfordPets   & 88.4 & \cellcolor{diag_green} 93.8 & 62.8 & 69.1 & 76.5 & 23.9 & 61.7 & 37.8 & 50.0 & 66.5 & 63.1 \\
StanfordCars &   88.9 & 93.8 & \cellcolor{diag_green}87.1 & 69.2 & 76.5 & 23.9 & 61.8 & 37.8 & 50.0 & 66.5 & 65.6 \\
Flowers102   & 89.6 & 93.8 & 87.1 & \cellcolor{diag_green}98.8 & 76.6 & 23.9 & 61.8 & 38.4 & 50.0 & 66.7 & 68.7  \\
Food101      & 89.7 & 93.8 & 87.1 & 98.8 & \cellcolor{diag_green}89.7 & 24.0 & 62.0 & 37.8 & 49.7 & 66.6 & 69.9 \\
FGVCAircraft & 94.9 & 93.8 & 87.1 & 98.8 & 89.7 & \cellcolor{diag_green}53.2 & 62.0 & 37.9 & 49.6 & 66.6 & 73.4 \\
SUN397       & 95.3 & 93.9 & 87.1 & 98.8 & 89.7 & 53.2 & \cellcolor{diag_green}76.0 & 38.9 & 49.9 & 66.7 & 75.0 \\
DTD          & 95.3 & 93.9 & 87.1 & 98.8 & 89.8 & 53.2 & 76.0 & \cellcolor{diag_green}75.5 & 48.7 & 66.7 & 78.5 \\
EuroSAT      & 95.3 & 93.9 & 87.1 & 98.8 & 89.8 & 53.2 & 76.0 & 75.5 & \cellcolor{diag_green}95.9 & 66.9 & 83.2 \\ 
UCF101       & \cellcolor{last_red}95.5 & \cellcolor{last_red}94.0 & \cellcolor{last_red}87.1 & \cellcolor{last_red}98.8 & \cellcolor{last_red}89.8 & \cellcolor{last_red}53.2 & \cellcolor{last_red}77.2 & \cellcolor{last_red}75.9 & \cellcolor{last_red}95.9 & \cellcolor{last_red}86.4 &   \textbf{85.4} \\
\midrule
\textbf{Average} &92.1 & 93.3 & 82.2 & 89.9 & 84.5 & 38.6 & 67.6 & 49.3 & 59.0 &  68.6 & \cellcolor{avg_yellow}\textbf{72.5} \\
\bottomrule
\end{tabular}
\end{adjustbox}
\end{table*}
\begin{table*}[!t]
\centering
\caption{\textbf{Comprehensive results of BETA in the black-box continual learning setting on the 16-shot task stream $\mathcal{S}$ with \textbf{CLIP-L/14-224} backbone.} Each row represents the performance on every dataset of the model trained after the corresponding task. We report accuracy (\%), where the diagonal (\colorbox{diag_green}{green}) represents the performance on the current task, and the last row (\colorbox{last_red}{red}) denotes the Last Accuracy.}
\begin{adjustbox}{width=0.7\textwidth}
\begin{tabular}{l cccccccccc >{\columncolor{avg_yellow}}c}
\toprule
& \rotatebox{90}{Caltech101} & \rotatebox{90}{OxfordPets} & \rotatebox{90}{StanfordCars} & \rotatebox{90}{Flowers102} & \rotatebox{90}{Food101} & \rotatebox{90}{FGVCAircraft} & \rotatebox{90}{SUN397} & \rotatebox{90}{DTD} & \rotatebox{90}{EuroSAT} & \rotatebox{90}{UCF101} & \textbf{Average} \\
\midrule
Zero-shot & 75.7 & 93.3 & 76.8 & 79.5 & 90.3 & 32.5 & 64.4 & 41.8 & 61.0 & 73.3 & 68.9 \\
\midrule
Caltech101   & \cellcolor{diag_green}87.1 & 93.4&  76.4&  78.6&  83.3 & 27.3 & 67.0 & 47.9 & 49.2  & 75.2 & 68.5 \\
OxfordPets   & 87.4 & \cellcolor{diag_green}94.8 & 76.4 & 78.6&  83.3&  27.3 & 67.0 & 47.8 & 48.0 & 75.1 & 68.6 \\
StanfordCars & 87.7  & 94.8 & \cellcolor{diag_green}89.2 & 78.6 & 83.3 & 27.3&  67.1 & 47.8 & 48.0 & 75.2 & 69.9 \\
Flowers102   & 89.2 & 94.8 & 89.2 & \cellcolor{diag_green}98.7 & 83.3 & 27.4 & 67.2 & 47.9 & 47.9 & 75.2 & 72.1 \\
Food101      & 89.2 & 94.8 & 89.2 & 98.7 & \cellcolor{diag_green}90.5 & 27.4 & 67.3  & 46.2 & 48.1 & 74.8 & 72.6 \\
FGVCAircraft & 93.2 & 94.8 & 89.2 & 98.7 & 90.5 & \cellcolor{diag_green}59.1 & 67.1 & 46.2 & 47.4 & 74.9 & 76.1 \\
SUN397       & 93.5 & 94.9 & 89.2 & 98.7 & 90.7 & 59.1 & \cellcolor{diag_green}76.2 & 46.7 & 49.1 & 74.9 & 77.3 \\
DTD          & 93.5 & 94.9 & 89.2 & 98.7 & 90.8 & 59.1 & 76.2 & \cellcolor{diag_green}71.2 & 49.3 & 74.7 & 79.8 \\
EuroSAT      & 93.5 &  94.9 & 89.2 & 98.7 & 90.8 & 59.1 & 76.2 & 71.2 & \cellcolor{diag_green}91.6 & 74.8 & 84.0 \\ 
UCF101       & \cellcolor{last_red}94.0 & \cellcolor{last_red}94.9 & \cellcolor{last_red}89.2 & \cellcolor{last_red}98.7 & \cellcolor{last_red}90.8 & \cellcolor{last_red}59.1 & \cellcolor{last_red}78.7 & \cellcolor{last_red}71.5 & \cellcolor{last_red}91.6 & \cellcolor{last_red}85.1 & \textbf{85.4} \\
\midrule
\textbf{Average} & 90.8 & 94.7 & 86.6 & 92.7 & 87.7 & 43.2 & 71.0 & 54.4 & 57.0 & 76.0  & \cellcolor{avg_yellow}\textbf{75.4} \\
\bottomrule
\end{tabular}
\end{adjustbox}
\end{table*}
\begin{table*}[!t]
\centering
\caption{\textbf{Comprehensive results of BETA in the black-box continual learning setting on the full-shot task stream $\mathcal{S}$ with \textbf{CLIP-L/14-224} backbone.} Each row represents the performance on every dataset of the model trained after the corresponding task. We report accuracy (\%), where the diagonal (\colorbox{diag_green}{green}) represents the performance on the current task, and the last row (\colorbox{last_red}{red}) denotes the Last Accuracy.}
\begin{adjustbox}{width=0.7\textwidth}
\begin{tabular}{l cccccccccc >{\columncolor{avg_yellow}}c}
\toprule
& \rotatebox{90}{Caltech101} & \rotatebox{90}{OxfordPets} & \rotatebox{90}{StanfordCars} & \rotatebox{90}{Flowers102} & \rotatebox{90}{Food101} & \rotatebox{90}{FGVCAircraft} & \rotatebox{90}{SUN397} & \rotatebox{90}{DTD} & \rotatebox{90}{EuroSAT} & \rotatebox{90}{UCF101} & \textbf{Average} \\
\midrule
Zero-shot & 75.7 & 93.3 & 76.8 & 79.5 & 90.3 & 32.5 & 64.4 & 41.8 & 61.0 & 73.3 & 68.9 \\
\midrule
Caltech101   & \cellcolor{diag_green}88.0 & 93.5&  76.4&  79.0&  83.6 & 30.4 & 67.1 & 47.8 & 49.3  & 75.2 & 69.0 \\
OxfordPets   & 88.2 & \cellcolor{diag_green}95.4 & 76.4 & 79.0&  83.6&  30.4 & 67.0 & 47.7 & 48.0 & 75.1 & 69.1 \\
StanfordCars & 88.6  & 95.4 & \cellcolor{diag_green}91.0 & 79.0 & 83.6 & 30.4&  67.2 & 47.7 & 48.0 & 75.1 & 70.6 \\
Flowers102   & 90.3 & 95.4 & 91.0 & \cellcolor{diag_green}99.2 & 83.6 & 30.4 & 67.2 & 47.9 & 48.0 & 75.1 & 72.8 \\
Food101      & 90.5 & 95.4 & 91.0 & 99.2 & \cellcolor{diag_green}93.1 & 30.4 & 67.4  & 46.7 & 48.1 & 74.7 & 73.7 \\
FGVCAircraft & 96.2 & 95.4 & 91.0 & 99.2 & 93.1 & \cellcolor{diag_green}63.2 & 67.3 & 46.5 & 48.2 & 74.7 & 77.5 \\
SUN397       & 96.6 & 95.5 & 90.9 & 99.2 & 93.1 & 63.2 & \cellcolor{diag_green}79.1 & 46.3 & 48.7 & 74.2 & 78.7 \\
DTD          & 96.6 & 95.4 & 90.9 & 99.2 & 93.1 & 63.2 & 79.1 & \cellcolor{diag_green}79.2 & 48.6 & 74.1 & 81.9 \\
EuroSAT      & 96.6 &  95.4 & 90.9 & 99.2 & 93.1 & 63.2 & 79.2 & 79.3 & \cellcolor{diag_green}97.2 & 74.4 & 86.9 \\ 
UCF101       & \cellcolor{last_red}96.8 & \cellcolor{last_red}95.5 & \cellcolor{last_red}90.9 & \cellcolor{last_red}99.2 & \cellcolor{last_red}93.1 & \cellcolor{last_red}63.2 & \cellcolor{last_red}80.8 & \cellcolor{last_red}79.4 & \cellcolor{last_red}97.2 & \cellcolor{last_red}90.0 & \textbf{88.6} \\
\midrule
\textbf{Average} &92.8 & 95.2 & 88.0 & 93.1 & 89.3 & 46.8 & 72.1 & 56.9 & 58.1 &  76.3 & \cellcolor{avg_yellow}\textbf{76.9} \\
\bottomrule
\end{tabular}
\end{adjustbox}
\end{table*}
\begin{table*}[!t]
\centering
\caption{\textbf{Comprehensive results of BETA in the black-box continual learning setting on the 16-shot task stream $\mathcal{S}$ with \textbf{SigLIP-B/16-224} backbone.} Each row represents the performance on every dataset of the model trained after the corresponding task. We report accuracy (\%), where the diagonal (\colorbox{diag_green}{green}) represents the performance on the current task, and the last row (\colorbox{last_red}{red}) denotes the Last Accuracy.}
\begin{adjustbox}{width=0.7\textwidth}
\begin{tabular}{l cccccccccc >{\columncolor{avg_yellow}}c}
\toprule
& \rotatebox{90}{Caltech101} & \rotatebox{90}{OxfordPets} & \rotatebox{90}{StanfordCars} & \rotatebox{90}{Flowers102} & \rotatebox{90}{Food101} & \rotatebox{90}{FGVCAircraft} & \rotatebox{90}{SUN397} & \rotatebox{90}{DTD} & \rotatebox{90}{EuroSAT} & \rotatebox{90}{UCF101} & \textbf{Average} \\
\midrule
Zero-shot & 85.3 & 92.8 & 89.0 & 81.7 & 88.0 & 29.4 & 65.7 & 55.1 & 31.6 & 72.2 &  69.1 \\
\midrule
Caltech101   & \cellcolor{diag_green}86.7 & 92.0&  89.2&  82.9&  69.9 & 28.9 & 67.7 & 63.9 & 19.8  & 73.1 & 67.4 \\
OxfordPets   & 86.6 & \cellcolor{diag_green}94.8 & 89.2 & 82.9&  70.0&  28.9 & 67.7 & 63.8 & 19.8 & 73.1 & 67.7 \\
StanfordCars & 86.5  & 94.8 & \cellcolor{diag_green}93.7 & 82.9 & 70.0 & 28.8&  67.8 & 63.8 & 19.8 & 73.1 & 68.1 \\
Flowers102   & 86.9 & 94.8 & 93.7 & \cellcolor{diag_green}99.2 & 70.0 & 28.8 & 67.7 & 63.9 & 19.8 & 73.1 & 69.8 \\
Food101      & 87.2 & 94.8 & 93.7 & 99.2 & \cellcolor{diag_green}89.0 & 28.8 & 68.1  & 62.1 & 19.7 & 73.2 & 71.6 \\
FGVCAircraft & 94.4 & 94.8 & 93.7 & 99.2 & 89.0 & \cellcolor{diag_green}53.4 & 68.1 & 61.6 & 19.7 & 73.2 & 74.7 \\
SUN397       & 95.2 & 94.9 & 93.6 & 99.2 & 89.3 & 53.5 & \cellcolor{diag_green}73.6 & 62.6 & 18.1 & 73.4 & 75.3 \\
DTD          & 95.2 & 94.9 & 93.6 & 99.2 & 89.4 & 53.5 & 73.7 & \cellcolor{diag_green}76.2 & 19.5 & 73.5 & 76.9 \\
EuroSAT      & 95.2 &  94.9 & 93.6 & 99.2 & 89.4 & 53.5 & 73.8 & 76.2 & \cellcolor{diag_green}88.0 & 73.8 & 83.8 \\ 
UCF101       & \cellcolor{last_red}95.4 & \cellcolor{last_red}95.0 & \cellcolor{last_red}93.6 & \cellcolor{last_red}99.2 & \cellcolor{last_red}89.4 & \cellcolor{last_red}53.5 & \cellcolor{last_red}76.4 & \cellcolor{last_red}76.7 & \cellcolor{last_red}87.9 & \cellcolor{last_red}80.8 & \textbf{84.8} \\
\midrule
\textbf{Average} & 90.9 & 94.6 & 82.8 & 94.3 & 81.5 & 41.2 & 70.5 & 67.1 & 33.2 & 74.0  & \cellcolor{avg_yellow}\textbf{74.0} \\
\bottomrule
\end{tabular}
\end{adjustbox}
\end{table*}
\begin{table*}[!t]
\centering
\caption{\textbf{Comprehensive results of BETA in the black-box continual learning setting on the full-shot task stream $\mathcal{S}$ with \textbf{SigLIP-B/16-224} backbone.} Each row represents the performance on every dataset of the model trained after the corresponding task. We report accuracy (\%), where the diagonal (\colorbox{diag_green}{green}) represents the performance on the current task, and the last row (\colorbox{last_red}{red}) denotes the Last Accuracy.}
\begin{adjustbox}{width=0.7\textwidth}
\begin{tabular}{l cccccccccc >{\columncolor{avg_yellow}}c}
\toprule
& \rotatebox{90}{Caltech101} & \rotatebox{90}{OxfordPets} & \rotatebox{90}{StanfordCars} & \rotatebox{90}{Flowers102} & \rotatebox{90}{Food101} & \rotatebox{90}{FGVCAircraft} & \rotatebox{90}{SUN397} & \rotatebox{90}{DTD} & \rotatebox{90}{EuroSAT} & \rotatebox{90}{UCF101} & \textbf{Average} \\
\midrule
Zero-shot & 85.3 & 92.8 & 89.0 & 81.7 & 88.0 & 29.4 & 65.7 & 55.1 & 31.6 & 72.2 &  69.1 \\
\midrule
Caltech101   & \cellcolor{diag_green}89.4 & 92.6&  89.3&  83.7&  70.8 & 28.9 & 67.6 & 64.1 & 19.8  & 73.1 & 67.9 \\
OxfordPets   & 89.3 & \cellcolor{diag_green}95.3 & 89.3 & 83.7&  70.8&  28.9 & 67.6 & 63.9 & 19.9 & 73.1 & 68.2 \\
StanfordCars & 89.3  & 95.3 & \cellcolor{diag_green}94.6 & 83.7 & 70.8 & 28.8&  67.7 & 63.9 & 19.8 & 73.2 & 68.7 \\
Flowers102   & 90.0 & 95.3 & 94.6 & \cellcolor{diag_green}99.6 & 70.9 & 28.8 & 67.7 & 63.9 & 19.8 & 73.1 & 70.4 \\
Food101      & 90.6 & 95.3 & 94.6 & 99.6 & \cellcolor{diag_green}91.3 & 28.8 & 68.1  & 62.5 & 19.7 & 73.1 & 72.4 \\
FGVCAircraft & 96.6 & 95.3 & 94.6 & 99.6 & 91.3 & \cellcolor{diag_green}58.1 & 68.0 & 61.8 & 19.7 & 73.1 & 75.8 \\
SUN397       & 97.4 & 95.3 & 94.6 & 99.6 & 91.4 & 58.1 & \cellcolor{diag_green}76.5 & 62.4 & 17.7 & 73.6 & 76.7 \\
DTD          & 97.4 & 95.3 & 94.6 & 99.6 & 91.4 & 58.1 & 76.6 & \cellcolor{diag_green}81.4 & 19.8 & 74.0 & 78.8 \\
EuroSAT      & 97.4  & 95.3 & 94.6 & 99.6 & 91.4 & 58.1 & 76.6 & 81.3 & \cellcolor{diag_green}96.8 & 74.3 & 86.5 \\ 
UCF101       & \cellcolor{last_red}97.4 & \cellcolor{last_red}95.3 & \cellcolor{last_red}94.6 & \cellcolor{last_red}99.6 & \cellcolor{last_red}91.5 & \cellcolor{last_red}58.1 & \cellcolor{last_red}78.5 & \cellcolor{last_red}81.4 & \cellcolor{last_red}96.7 & \cellcolor{last_red}86.4 & \textbf{88.0} \\
\midrule
\textbf{Average} &93.5 & 95.0 & 93.5 & 94.8 & 83.2 & 43.5 & 71.5 & 68.7 & 35.0 &  74.7 & \cellcolor{avg_yellow}\textbf{75.3} \\
\bottomrule
\end{tabular}
\end{adjustbox}
\end{table*}
\begin{table*}[!t]
\centering
\caption{\textbf{Comprehensive results of BETA in the black-box continual learning setting on the 16-shot task stream $\mathcal{S}$ with \textbf{SigLIP-So/14-224} backbone.} Each row represents the performance on every dataset of the model trained after the corresponding task. We report accuracy (\%), where the diagonal (\colorbox{diag_green}{green}) represents the performance on the current task, and the last row (\colorbox{last_red}{red}) denotes the Last Accuracy.}
\begin{adjustbox}{width=0.7\textwidth}
\begin{tabular}{l cccccccccc >{\columncolor{avg_yellow}}c}
\toprule
& \rotatebox{90}{Caltech101} & \rotatebox{90}{OxfordPets} & \rotatebox{90}{StanfordCars} & \rotatebox{90}{Flowers102} & \rotatebox{90}{Food101} & \rotatebox{90}{FGVCAircraft} & \rotatebox{90}{SUN397} & \rotatebox{90}{DTD} & \rotatebox{90}{EuroSAT} & \rotatebox{90}{UCF101} & \textbf{Average} \\
\midrule
Zero-shot & 85.4 & 95.0 & 87.5 & 89.5 & 92.6 & 50.1 & 72.0 & 56.4 & 49.4 & 77.6 & 75.5\\
\midrule
Caltech101   & \cellcolor{diag_green}87.0 & 95.4&  87.9&  92.0&  83.9 & 50.6 & 74.4 & 62.2 & 27.4  & 80.8 & 74.2 \\
OxfordPets   & 87.0 & \cellcolor{diag_green}96.2 & 87.9 & 92.1&  83.8&  50.6 & 74.4 & 62.2 & 27.4 & 80.8 & 74.2 \\
StanfordCars & 87.1  & 96.2 & \cellcolor{diag_green}95.3 & 92.1 & 83.9 & 50.6&  74.3 & 62.2 & 27.4 & 80.9 & 75.0 \\
Flowers102   & 88.9 & 96.2 & 95.3 & \cellcolor{diag_green}99.6 & 83.9 & 50.6 & 74.3 & 62.6 & 27.4 & 80.8 & 76.0 \\
Food101      & 89.2 & 96.2 & 95.3 & 99.6 & \cellcolor{diag_green}92.7 & 50.6 & 74.6  & 63.0 & 27.3 & 80.9 & 76.9 \\
FGVCAircraft & 94.1 & 96.2 & 95.3 & 99.6 & 92.7 & \cellcolor{diag_green}69.0 & 74.6 & 63.0 & 27.3 & 80.8 & 79.3 \\
SUN397       & 94.6 & 96.2 & 95.3 & 99.6 & 93.0 & 69.0 & \cellcolor{diag_green}78.8 & 63.0 & 27.5 & 80.2 & 79.7 \\
DTD          & 94.6 & 96.2 & 95.3 & 99.6 & 93.0 & 69.0 & 79.0 & \cellcolor{diag_green}78.7 & 28.9 & 80.7 & 81.5 \\
EuroSAT      & 94.6  & 96.2 & 95.3 & 99.6 & 93.0 & 69.0 & 79.1 & 78.7 & \cellcolor{diag_green}90.1 & 80.9 & 87.7 \\ 
UCF101       & \cellcolor{last_red}94.6 & \cellcolor{last_red}96.2 & \cellcolor{last_red}95.3 & \cellcolor{last_red}99.6 & \cellcolor{last_red}93.0 & \cellcolor{last_red}69.0 & \cellcolor{last_red}81.0 & \cellcolor{last_red}79.0 & \cellcolor{last_red}90.1 & \cellcolor{last_red}82.7 & \textbf{88.1} \\
\midrule
\textbf{Average} & 91.2 & 96.1 & 93.8 & 97.3 & 89.3 & 59.8 & 76.5 & 67.5 & 40.1 & 81.0  & \cellcolor{avg_yellow}\textbf{79.3} \\
\bottomrule
\end{tabular}
\end{adjustbox}
\end{table*}
\begin{table*}[!t]
\centering
\caption{\textbf{Comprehensive results of BETA in the black-box continual learning setting on the full-shot task stream $\mathcal{S}$ with \textbf{SigLIP-So/14-224} backbone.} Each row represents the performance on every dataset of the model trained after the corresponding task. We report accuracy (\%), where the diagonal (\colorbox{diag_green}{green}) represents the performance on the current task, and the last row (\colorbox{last_red}{red}) denotes the Last Accuracy.}
\begin{adjustbox}{width=0.7\textwidth}
\begin{tabular}{l cccccccccc >{\columncolor{avg_yellow}}c}
\toprule
& \rotatebox{90}{Caltech101} & \rotatebox{90}{OxfordPets} & \rotatebox{90}{StanfordCars} & \rotatebox{90}{Flowers102} & \rotatebox{90}{Food101} & \rotatebox{90}{FGVCAircraft} & \rotatebox{90}{SUN397} & \rotatebox{90}{DTD} & \rotatebox{90}{EuroSAT} & \rotatebox{90}{UCF101} & \textbf{Average} \\
\midrule
Zero-shot & 85.4 & 95.0 & 87.5 & 89.5 & 92.6 & 50.1 & 72.0 & 56.4 & 49.4 & 77.6 & 75.5\\
\midrule
Caltech101   & \cellcolor{diag_green}88.3 & 95.5&  87.9&  92.7&  84.0 & 50.5 & 74.4 & 62.2 & 27.4  & 80.8 & 74.4 \\
OxfordPets   & 88.4 & \cellcolor{diag_green}96.4 & 87.9 & 92.7&  84.1&  50.5 & 74.3 & 62.2 & 27.4 & 80.8 & 74.5 \\
StanfordCars & 88.6  & 96.4 & \cellcolor{diag_green}95.9 & 92.7 & 83.9 & 50.5&  74.3 & 62.2 & 27.4 & 80.9 & 75.3 \\
Flowers102   & 90.5 & 96.4 & 95.9 & \cellcolor{diag_green}99.8 & 84.0 & 50.5 & 74.3 & 62.7 & 27.4 & 80.9 & 76.2 \\
Food101      & 91.0 & 96.4 & 95.9 & 99.8 & \cellcolor{diag_green}94.5 & 50.5 & 74.7  & 62.9 & 27.3 & 80.8 & 77.4 \\
FGVCAircraft & 96.1 & 96.4 & 95.9 & 99.8 & 94.5 & \cellcolor{diag_green}71.7 & 74.6 & 62.9 & 27.3 & 80.9 & 80.0 \\
SUN397       & 96.4 & 96.4 & 95.9 & 99.8 & 94.6 & 71.7 & \cellcolor{diag_green}81.0 & 63.2 & 27.7 & 80.7 & 80.7 \\
DTD          & 96.4 & 96.4 & 95.9 & 99.8 & 94.7 & 71.7 & 81.1 & \cellcolor{diag_green}83.5 & 28.4 & 80.7 & 82.9 \\
EuroSAT      & 96.4  & 96.4 & 95.9 & 99.8 & 94.7 & 71.7 & 81.1 & 83.5 & \cellcolor{diag_green}97.1 & 81.0 & 89.8 \\ 
UCF101       & \cellcolor{last_red}96.4 & \cellcolor{last_red}96.4 & \cellcolor{last_red}95.9 & \cellcolor{last_red}99.8 & \cellcolor{last_red}94.7 & \cellcolor{last_red}71.7 & \cellcolor{last_red}82.9 & \cellcolor{last_red}83.5 & \cellcolor{last_red}97.1 & \cellcolor{last_red}90.8 & \textbf{90.9} \\
\midrule
\textbf{Average} &92.8 & 96.3 & 94.3 & 97.7 & 90.4 & 61.1 & 77.3 & 68.9 & 41.5 &  81.8 & \cellcolor{avg_yellow}\textbf{80.2} \\
\bottomrule
\end{tabular}
\end{adjustbox}
\end{table*}
\begin{table*}[!t]
\centering
\caption{\textbf{Comprehensive results of BETA in the black-box continual learning setting on the 16-shot task stream $\mathcal{S}$ with \textbf{SigLIP2-B/16-224} backbone.} Each row represents the performance on every dataset of the model trained after the corresponding task. We report accuracy (\%), where the diagonal (\colorbox{diag_green}{green}) represents the performance on the current task, and the last row (\colorbox{last_red}{red}) denotes the Last Accuracy.}
\begin{adjustbox}{width=0.7\textwidth}
\begin{tabular}{l cccccccccc >{\columncolor{avg_yellow}}c}
\toprule
& \rotatebox{90}{Caltech101} & \rotatebox{90}{OxfordPets} & \rotatebox{90}{StanfordCars} & \rotatebox{90}{Flowers102} & \rotatebox{90}{Food101} & \rotatebox{90}{FGVCAircraft} & \rotatebox{90}{SUN397} & \rotatebox{90}{DTD} & \rotatebox{90}{EuroSAT} & \rotatebox{90}{UCF101} & \textbf{Average} \\
\midrule
Zero-shot & 85.8 & 92.7 & 92.5 & 81.6 & 88.8 & 28.0 & 69.4 & 55.1 & 35.8 & 66.2 & 69.6 \\
\midrule
Caltech101   & \cellcolor{diag_green}92.1 & 92.4&  92.4&  80.0&  73.3 & 27.4 & 70.2 & 60.4 & 32.6  & 71.3 & 69.2 \\
OxfordPets   & 92.0 & \cellcolor{diag_green}95.0 & 92.4 & 80.0&  73.3&  27.4 & 70.2 & 60.3 & 32.6 & 71.3 & 69.5 \\
StanfordCars & 92.0  & 95.0 & \cellcolor{diag_green}94.0 & 80.0 & 73.3 & 27.4&  70.1 & 60.5 & 32.6 & 71.3 & 69.6 \\
Flowers102   & 92.0 & 95.0 & 94.0 & \cellcolor{diag_green}99.4 & 73.3 & 27.4 & 70.1 & 60.5 & 32.6 & 71.4 & 71.6 \\
Food101      & 92.0 & 95.0 & 94.0 & 99.4 & \cellcolor{diag_green}89.7 & 27.7 & 70.8  & 60.4 & 33.4 & 71.1 & 73.4 \\
FGVCAircraft & 95.2 & 95.0 & 94.0 & 99.4 & 89.7 & \cellcolor{diag_green}60.3 & 70.5 & 60.5 & 33.4 & 71.2 & 76.9 \\
SUN397       & 95.2 & 95.0 & 94.0 & 99.4 & 90.1 & 60.3 & \cellcolor{diag_green}76.4 & 61.5 & 30.7 & 71.2 & 77.4 \\
DTD          & 95.2 & 95.0 & 94.0 & 99.4 & 90.2 & 60.3 & 76.4 & \cellcolor{diag_green}76.7 & 34.0 & 71.1 & 79.2 \\
EuroSAT      & 95.2  & 95.0 & 94.0 & 99.4 & 90.2 & 60.3 & 76.5 & 76.7 & \cellcolor{diag_green}87.9 & 71.6 & 84.7 \\ 
UCF101       & \cellcolor{last_red}95.2 & \cellcolor{last_red}95.0 & \cellcolor{last_red}94.0 & \cellcolor{last_red}99.4 & \cellcolor{last_red}90.2 & \cellcolor{last_red}60.3 & \cellcolor{last_red}78.0 & \cellcolor{last_red}76.8 & \cellcolor{last_red}87.9 & \cellcolor{last_red}83.0 & \textbf{86.0} \\
\midrule
\textbf{Average} & 93.6 & 94.7 & 93.7 & 93.6 & 83.3 & 43.9 & 72.9 & 65.4 & 43.8 & 72.5  & \cellcolor{avg_yellow}\textbf{75.7} \\
\bottomrule
\end{tabular}
\end{adjustbox}
\end{table*}
\begin{table*}[!t]
\centering
\caption{\textbf{Comprehensive results of BETA in the black-box continual learning setting on the full-shot task stream $\mathcal{S}$ with \textbf{SigLIP2-B/16-224} backbone.} Each row represents the performance on every dataset of the model trained after the corresponding task. We report accuracy (\%), where the diagonal (\colorbox{diag_green}{green}) represents the performance on the current task, and the last row (\colorbox{last_red}{red}) denotes the Last Accuracy.}
\begin{adjustbox}{width=0.7\textwidth}
\begin{tabular}{l cccccccccc >{\columncolor{avg_yellow}}c}
\toprule
& \rotatebox{90}{Caltech101} & \rotatebox{90}{OxfordPets} & \rotatebox{90}{StanfordCars} & \rotatebox{90}{Flowers102} & \rotatebox{90}{Food101} & \rotatebox{90}{FGVCAircraft} & \rotatebox{90}{SUN397} & \rotatebox{90}{DTD} & \rotatebox{90}{EuroSAT} & \rotatebox{90}{UCF101} & \textbf{Average} \\
\midrule
Zero-shot & 85.8 & 92.7 & 92.5 & 81.6 & 88.8 & 28.0 & 69.4 & 55.1 & 35.8 & 66.2 & 69.6 \\
\midrule
Caltech101   & \cellcolor{diag_green}93.8 & 93.6&  92.5&  80.8&  74.0 & 27.6 & 70.2 & 60.7 & 32.6  & 71.5 & 69.7 \\
OxfordPets   & 93.8 & \cellcolor{diag_green}95.6 & 92.5 & 80.8&  73.9&  27.6 & 70.2 & 60.6 & 32.6 & 71.6 & 69.9 \\
StanfordCars & 93.8  & 95.6 & \cellcolor{diag_green}94.4 & 80.8 & 74.1 & 27.6&  70.1 & 60.5 & 32.6 & 71.6 & 70.1 \\
Flowers102   & 94.0 & 95.6 & 94.4 & \cellcolor{diag_green}99.6 & 74.1 & 27.6 & 70.0 & 60.6 & 32.6 & 71.5 & 72.0 \\
Food101      & 94.0 & 95.6 & 94.4 & 99.6 & \cellcolor{diag_green}92.1 & 27.7 & 70.7  & 60.4 & 33.2 & 70.8 & 73.9 \\
FGVCAircraft & 97.3 & 95.6 & 94.4 & 99.6 & 92.1 & \cellcolor{diag_green}65.6 & 70.4 & 60.5 & 33.2 & 71.0 & 78.0 \\
SUN397       & 97.4 & 95.5 & 94.4 & 99.6 & 92.3 & 65.6 & \cellcolor{diag_green}79.2 & 61.4 & 30.2 & 69.8 & 78.5 \\
DTD          & 97.4 & 95.5 & 94.4 & 99.6 & 92.3 & 65.6 & 79.2 & \cellcolor{diag_green}82.2 & 31.9 & 70.0 & 80.8 \\
EuroSAT      & 97.4  & 95.5 & 94.4 & 99.6 & 92.3 & 65.6 & 79.2 & 82.1 & \cellcolor{diag_green}96.6 & 70.3 & 87.3 \\ 
UCF101       & \cellcolor{last_red}97.4 & \cellcolor{last_red}95.5 & \cellcolor{last_red}94.4 & \cellcolor{last_red}99.6 & \cellcolor{last_red}92.3 & \cellcolor{last_red}65.6 & \cellcolor{last_red}80.0 & \cellcolor{last_red}82.1 & \cellcolor{last_red}96.6 & \cellcolor{last_red}88.0 & \textbf{89.2} \\
\midrule
\textbf{Average} &95.6 & 95.4 & 94.0 & 94.0 & 84.9 & 46.6 & 73.9 & 67.1 & 45.2 &  72.6 & \cellcolor{avg_yellow}\textbf{76.9} \\
\bottomrule
\end{tabular}
\end{adjustbox}
\end{table*}
\begin{table*}[!t]
\centering
\caption{\textbf{Comprehensive results of BETA in the black-box continual learning setting on the 16-shot task stream $\mathcal{S}$ with \textbf{SigLIP2-So/14-224} backbone.} Each row represents the performance on every dataset of the model trained after the corresponding task. We report accuracy (\%), where the diagonal (\colorbox{diag_green}{green}) represents the performance on the current task, and the last row (\colorbox{last_red}{red}) denotes the Last Accuracy.}
\begin{adjustbox}{width=0.7\textwidth}
\begin{tabular}{l cccccccccc >{\columncolor{avg_yellow}}c}
\toprule
& \rotatebox{90}{Caltech101} & \rotatebox{90}{OxfordPets} & \rotatebox{90}{StanfordCars} & \rotatebox{90}{Flowers102} & \rotatebox{90}{Food101} & \rotatebox{90}{FGVCAircraft} & \rotatebox{90}{SUN397} & \rotatebox{90}{DTD} & \rotatebox{90}{EuroSAT} & \rotatebox{90}{UCF101} & \textbf{Average} \\
\midrule
Zero-shot & 85.3 & 94.2 & 95.0 & 89.2 & 92.5 & 55.0 & 70.8 & 60.1 & 43.3 & 76.0 & 76.1 \\
\midrule
Caltech101   & \cellcolor{diag_green}90.6 & 95.2&  94.9&  89.5&  82.0 & 55.2 & 72.6 & 64.1 & 40.7  & 80.4 & 76.5 \\
OxfordPets   & 90.7 & \cellcolor{diag_green}96.2 & 94.9 & 89.5&  81.7&  55.2 & 72.6 & 63.9 & 40.7 & 80.5 & 76.6 \\
StanfordCars & 90.6  & 96.2 & \cellcolor{diag_green}95.9 & 89.5 & 82.0 & 55.2&  72.5 & 63.8 & 40.7 & 80.4 & 76.7 \\
Flowers102   & 91.0 & 96.2 & 95.9 & \cellcolor{diag_green}99.8 & 82.0 & 55.2 & 72.5 & 63.9 & 40.7 & 80.4 & 77.8 \\
Food101      & 91.0 & 96.2 & 95.9 & 99.8 & \cellcolor{diag_green}93.0 & 55.3 & 73.3  & 64.7 & 41.1 & 80.4 & 79.1 \\
FGVCAircraft & 95.5 & 96.2 & 95.9 & 99.8 & 93.0 & \cellcolor{diag_green}77.3 & 73.3 & 64.7 & 41.1 & 80.4 & 81.7 \\
SUN397       & 95.8 & 96.2 & 95.9 & 99.8 & 93.3 & 77.3 & \cellcolor{diag_green}79.4 & 65.4 & 40.9 & 79.6 & 82.4 \\
DTD          & 95.8 & 96.2 & 95.9 & 99.8 & 93.3 & 77.3 & 79.5 & \cellcolor{diag_green}78.3 & 43.5 & 79.7 & 83.9 \\
EuroSAT      & 95.7 &  96.2 & 95.9 & 99.8 & 93.3 & 77.3 & 79.5 & 78.4 & \cellcolor{diag_green}86.9 & 80.0 & 88.3 \\ 
UCF101       & \cellcolor{last_red}95.7 & \cellcolor{last_red}96.2 & \cellcolor{last_red}95.9 & \cellcolor{last_red}99.8 & \cellcolor{last_red}93.4 & \cellcolor{last_red}77.3 & \cellcolor{last_red}81.1 & \cellcolor{last_red}78.5 & \cellcolor{last_red}87.0 & \cellcolor{last_red}84.4 & \textbf{88.9} \\
\midrule
\textbf{Average} & 93.2 & 96.1 & 95.7 & 96.7 & 88.7 & 66.3 & 75.6 & 68.6 & 50.3 & 80.6  & \cellcolor{avg_yellow}\textbf{81.2} \\
\bottomrule
\end{tabular}
\end{adjustbox}
\end{table*}
\begin{table*}[!t]
\centering
\caption{\textbf{Comprehensive results of BETA in the black-box continual learning setting on the full-shot task stream $\mathcal{S}$ with \textbf{SigLIP2-So/14-224} backbone.} Each row represents the performance on every dataset of the model trained after the corresponding task. We report accuracy (\%), where the diagonal (\colorbox{diag_green}{green}) represents the performance on the current task, and the last row (\colorbox{last_red}{red}) denotes the Last Accuracy.}
\begin{adjustbox}{width=0.7\textwidth}
\begin{tabular}{l cccccccccc >{\columncolor{avg_yellow}}c}
\toprule
& \rotatebox{90}{Caltech101} & \rotatebox{90}{OxfordPets} & \rotatebox{90}{StanfordCars} & \rotatebox{90}{Flowers102} & \rotatebox{90}{Food101} & \rotatebox{90}{FGVCAircraft} & \rotatebox{90}{SUN397} & \rotatebox{90}{DTD} & \rotatebox{90}{EuroSAT} & \rotatebox{90}{UCF101} & \textbf{Average} \\
\midrule
Zero-shot & 85.3 & 94.2 & 95.0 & 89.2 & 92.5 & 55.0 & 70.8 & 60.1 & 43.3 & 76.0 & 76.1 \\
\midrule
Caltech101   & \cellcolor{diag_green}92.8 & 95.1&  94.9&  89.4&  82.0 & 55.2 & 72.6 & 63.8 & 40.7  & 80.5 & 76.7 \\
OxfordPets   & 92.9 & \cellcolor{diag_green}96.5 & 94.9 & 89.4&  82.0&  55.2 & 72.5 & 63.8 & 40.7 & 80.4 & 76.8 \\
StanfordCars & 92.8  & 96.5 & \cellcolor{diag_green}96.3 & 89.4 & 82.0 & 55.2&  72.4 & 63.8 & 40.7 & 80.4 & 77.0 \\
Flowers102   & 93.3 & 96.5 & 96.3 & \cellcolor{diag_green}99.9 & 82.0 & 55.2 & 72.4 & 63.8 & 40.7 & 80.4 & 78.1 \\
Food101      & 93.3 & 96.5 & 96.3 & 99.9 & \cellcolor{diag_green}95.1 & 55.3 & 73.6  & 64.5 & 41.1 & 80.2 & 79.6 \\
FGVCAircraft & 97.0 & 96.5 & 96.3 & 99.9 & 95.1 & \cellcolor{diag_green}79.6 & 73.5 & 64.8 & 41.1 & 80.2 & 82.4 \\
SUN397       & 97.2 & 96.4 & 96.3 & 99.9 & 95.2 & 79.6 & \cellcolor{diag_green}82.1 & 65.1 & 40.9 & 79.1 & 83.2 \\
DTD          & 97.2 & 96.4 & 96.3 & 99.9 & 95.2 & 79.6 & 82.1 & \cellcolor{diag_green}84.3 & 41.6 & 79.0 & 85.2 \\
EuroSAT      & 97.2 &  96.4 & 96.3 & 99.9 & 95.2 & 79.6 & 82.1 & 84.3 & \cellcolor{diag_green}96.4 & 79.4 & 90.7 \\ 
UCF101       & \cellcolor{last_red}97.2 & \cellcolor{last_red}96.4 & \cellcolor{last_red}96.3 & \cellcolor{last_red}99.9 & \cellcolor{last_red}95.2 & \cellcolor{last_red}79.6 & \cellcolor{last_red}83.3 & \cellcolor{last_red}84.4 & \cellcolor{last_red}96.4 & \cellcolor{last_red}92.7 & \textbf{92.1} \\
\midrule
\textbf{Average} & 95.1 & 96.3 & 96.0 & 96.7 & 89.9 & 67.4 & 76.7 & 70.3 & 52.0 & 81.2  & \cellcolor{avg_yellow}\textbf{82.2} \\
\bottomrule
\end{tabular}
\end{adjustbox}
\end{table*}
\end{document}